\documentclass[a4paper, 11pt]{article}

\usepackage{amsmath, amsfonts, amssymb}
\usepackage{graphicx}
\usepackage{subfigure}
\usepackage{hyperref}
\usepackage{url}
\usepackage{xcolor}
\usepackage{booktabs}
\usepackage{multirow}
\usepackage{makecell}
\usepackage{nicefrac}
\usepackage{microtype}
\usepackage{authblk}
\usepackage{cite}
\newcommand{\modified}[1]{\textcolor{blue}{#1}}
\renewcommand{\modified}[1]{#1}

\graphicspath{{./images/}}

\title{Inverse Evolution Data Augmentation for Neural PDE Solvers}

\author{    
    Chaoyu Liu$^{1}$, Chris Budd$^{2}$, and Carola-Bibiane Schönlieb$^{1}$ \\
    \small $^1$Department of Applied Mathematical and Theoretical Physics, University of Cambridge \\
    \small $^2$Department of Mathematical Sciences, University of Bath \\
    \texttt{cl920@cam.ac.uk}
}

\date{}

\begin{document}

\maketitle

\begin{abstract}
    Neural networks have emerged as promising tools for solving partial differential equations (PDEs), particularly through the application of neural operators. Training neural operators typically requires a large amount of training data to ensure accuracy and generalization. In this paper, we propose a novel data augmentation method specifically designed for training neural operators on evolution equations. Our approach utilizes insights from inverse processes of these equations to efficiently generate data from random initialization that are combined with original data. To further enhance the accuracy of the augmented data, we introduce high-order inverse evolution schemes. These schemes consist of only a few explicit computation steps, yet the resulting data pairs can be proven to satisfy the corresponding implicit numerical schemes. In contrast to traditional PDE solvers that require small time steps or implicit schemes to guarantee accuracy, our data augmentation method employs explicit schemes with relatively large time steps, thereby significantly reducing computational costs. Accuracy and efficacy experiments confirm the effectiveness of our approach. Additionally, we validate our approach through experiments with the Fourier Neural Operator and UNet on three common evolution equations that are Burgers' equation, the Allen-Cahn equation and the Navier-Stokes equation. The results demonstrate a significant improvement in the performance and robustness of the Fourier Neural Operator when coupled with our inverse evolution data augmentation method.
\end{abstract}

\vspace{1em}


\vspace{0.5em}

\noindent \textbf{Subject Areas:} 68T07, 35Q35, 65N99.

\section{Introduction}

Partial differential equations (PDEs) play a fundamental role in modeling various phenomena across diverse fields, including molecular dynamics, fluid dynamics, weather forecasting, and astronomical simulations \cite{batchelor1967introduction, courant1967partial, evans2022partial, lelievre2016partial}. Many physical phenomena are inherently governed by complex PDEs. However, obtaining analytical solutions for these PDEs is often impractical. Consequently, numerical methods have become the primary means for simulating such systems. Despite their theoretical soundness and accuracy, conventional numerical methods often suffer from high computational costs and require specialized designs tailored to different classes of PDEs \cite{quarteroni2009numerical}. These challenges are particularly pronounced when addressing high-dimensional problems requiring fine discretization. 

In recent years, there has been a surge in the development of PDE solvers based on deep neural networks, aiming to create computationally efficient yet sufficiently accurate surrogates to classical solvers \cite{lu2021learning,raissi2019physics,xu2023physics}. One notable advancement in this area are the physics-informed neural networks \cite{cuomo2022scientific,raissi2019physics}. These networks compute differential operators using automatic differentiation and enforce equations and boundary conditions through residual loss functions. This method allows for mesh-free computation of differential operators and is advantageous for high-dimensional data. However, using automatic differentiation increases computational memory requirements, and the unsupervised nature of these methods can complicate the training process and impede convergence, especially in multi-scale dynamical systems \cite{lu2021deepxde}. 

Another promising approach is the neural operator framework, which learns mappings between function spaces \cite{bartolucci2024representation,li2021fourier,lu2021learning}. The theoretical foundation of neural operators lies in their ability to serve as universal approximators for any continuous operator \cite{kovachki2021universal,kovachki2023neural}. Compared to physics-informed methods, neural operators are easier to train and do not require automatic differentiation. However, the effectiveness of these methods relies heavily on the availability of substantial high-quality training data. In many instances, acquiring this data can be prohibitively expensive, difficult to obtain, or limited to low-resolution formats \cite{hersbach2020era5}. 

This reliance on high-quality training data is a fundamental limitation of neural operator methods. Some have proposed integrating physics-informed techniques into the training of neural operators to solve this problem\cite{wang2021learning,li2021physics}. However, these approaches may increase memory usage and computational costs due to the introduction of automatic differentiation and do not fundamentally resolve the data scarcity problem. Given this context, efficient data augmentation becomes a crucial task. While there is some literature on augmentation methods for neural operators, most focus on finding feasible transformations of solutions to generate new data \cite{brandstetter2022lie,fanaskov2023general,li2022physics}. However, since most transformations are linear or coordinate-dependent, they are incapable of generating data that exhibit adequate nonlinear variations from the original data. In this paper, we propose a novel approach to augment datasets for neural PDE solvers that does not rely on linear or coordinate-based transformations, and significantly expands the current data augmentation strategies.

\modified{There are many different types of PDEs. In physics, various evolutionary processes can be effectively modeled using evolution equations. In this paper, we focus on solving evolution equations, which can be generalized to the following form
\begin{equation}
\label{forward_pde}
\begin{aligned}
    u_t &= \mathcal{F}(u) & &\text{in } \Omega_T:= [0,T]\times\Omega\\
    u(0, x) &= g(x) & &\text{in } \Omega
\end{aligned}
\end{equation}
with periodic boundary condition. Here $\Omega$ represents an interval or a rectangular domain, $u = u(t, x)$ denotes the analytical solution of an evolution, $g(x)$ is the initial value and $\mathcal{F}$ refers to a combination of linear and nonlinear operators which can include a variety of gradient operators and nonlinear terms. For instance, $\mathcal{F}$ in the heat equation is a linear Laplacian operator, while for equations like the Allen-Cahn and Navier-Stokes equations, it involves gradient operators and nonlinear terms. For simplicity, we adopt periodic boundary conditions for equation \eqref{forward_pde}. Other boundary conditions can also be considered without loss of generality.}

These equations are time-dependent and frequently arise in a wide range of applications, such as fluid dynamics and weather forecasting. Such PDEs can be effectively addressed by neural operators utilizing recurrent neural network (RNN) structures similar to the Fourier Neural Operator (FNO) in 2D \cite{li2021fourier}. While neural operators have demonstrated proficiency in solving these PDEs, their heavy reliance on data is a limitation to their use. We propose a novel data augmentation to address this challenge. The primary motivation behind our method is that explicitly computing the inverse evolution is equivalent to implicitly computing the forward evolution, as explained in section \ref{data generation}. Therefore, our proposed data generation method can produce solutions consistent with reliable implicit schemes by performing only a few steps of explicit computation, which substantially alleviates the training data pressure for neural operators. In addition, in contrast to the original data characterized by predominantly smooth and stable solutions, our inverse evolution augmentation method produces twinned data sets with higher-frequency components. This augmentation significantly expands the solution space beyond that of the original data, thereby enhancing the resilience and robustness of the neural operators.

The main contributions of this work are:
\begin{quote}
\noindent
1. We propose a novel data generation method for neural operator-based PDE solvers based on inverse evolution. Experiments demonstrate the accuracy of the generated data, its distinctiveness from the training data, and its efficacy when applied to neural operators.

\noindent
2. We derive high-order formulas for our data generation method. Experimental results show that utilizing these high-order formulas significantly enhances the accuracy of the augmented data.

\noindent
3. We introduce a preprocessing method to address the instability issue that arises when handling differential equations with sharp interfaces. Experiments indicate that these techniques improve the accuracy of the generated data and allow for larger time intervals in the inverse evolution.
\end{quote}
\section{Data Generation Based on Inverse Evolution}
\label{data generation}

\paragraph{Neural Operator Training}\modified{A common approach to train a neural operator to solve evolution equations is by predicting the solution at a later time $t + t_0$ based on the solution at time $t$. Here $t_0$ is a fixed positive number. We denote a neural operator as $\mathcal{N}_\theta$, where $\theta$ represents the learnable parameters. Given a dataset of paired states, $\{(U^{k}, U^{k+1})\}_{k=1}^N$, where $U^k$ and $U^{k+1}$ are exact solutions with a time difference of $t_0$, the training objective for the neural operator in this setting is to minimize the following loss function with respect to $\theta$:}

\begin{equation}
    \theta^* = \arg \min_{\theta} \sum_{k=1}^{N} \left\| \mathcal{N}_\theta(U^{k}) - U^{k+1} \right\|_{L^2},
\end{equation}
\modified{where the $L^2$ norm measures the discrepancy between the predicted and true states. This objective ensures that the neural operator learns to approximate the transition between successive states in the dataset accurately. In practice, other norms can also be used to measure this discrepancy, such as the $H^2$ norm, depending on the specific requirements of the application.}

Longer trajectory predictions can be achieved by recursively feeding the predictions back into the neural operator. Consequently, with sufficient training pairs of the form $(u(t), u(t + t_0))$, one can train a neural operator to recursively solve the evolution equation.
It is important to note that one can also directly train the neural operator to give a solution for large $t_0$. However, in practice, the accuracy of the predictions diminishes for larger $t_0$, while autoregressive approaches with appropriate $t_0$ are found to perform substantially better \cite{gupta2022towards,li2022learning,wang2023long}.

For the neural operators, data plays a very important role. However, generating sufficient training data can be time-consuming. For evolution equations of the form of \eqref{forward_pde}, various numerical schemes have been designed to obtain accurate numerical solutions in regular domains \cite{eriksson1996computational,leveque2007finite,smith1985numerical}. One commonly used scheme for temporal discretization is the Forward Euler method. Despite its simplicity, the explicit Forward Euler  is prone to instability, necessitating very small time steps for stability and accuracy \cite{smith1985numerical}. Typically, the time step size for explicit numerical methods needs to be smaller than $O(h^2)$ for second-order differential equations and even smaller for high-order explicit numerical schemes, where $h$ is the mesh size of spatial discretization. Consequently, explicit methods demand significant computational time to find $u(t + t_0)$ given $u(t)$. To mitigate this, implicit methods are employed to avoid the small time steps. Although implicit methods offer high stability and accuracy, they often involve solving a nonlinear equation system at each iteration, which introduces additional computational iterations and implementation challenges for parallel computation. Semi-implicit methods can also be used to generate data, but they may smooth sharp interfaces in solutions and lose accuracy when their time step is relatively large.

To address this challenge, we propose a novel data augmentation method for evolution equations inspired by inverse evolution. Instead of solving for $u(t + t_0)$ from $u(t)$, our approach begins with $u(t + t_0)$ and seeks to determine the corresponding $u(t)$. Our method efficiently finds feasible values of $u(t)$ without solving any nonlinear equations. It also significantly increases the richness of the solution pairs which allows a more effective training of the neural operator. Additionally, it can be proved that the generated data pairs $(u(t), u(t + t_0))$ satisfy the form of the implicit schemes.
\paragraph{Inverse Evolution}
The concept of inverse evolution is introduced in \cite{liu2023inverse}, where inverse evolution serves as an error amplifier to regularize the output of neural networks. Essentially, inverse evolution is the inverse process of the original evolution. In our work, we employ inverse evolution to generated reliable data for the neural operators. The inverse evolution corresponding to the Eq. \eqref{forward_pde} follows the PDEs
\begin{equation}
\label{inverse_pde}
\begin{aligned}
    u_t &= -\mathcal{F}(u) & &\text{in } \Omega_T:= [0,T]\times\Omega\\
    u(0, x) &= g(x) & &\text{in } \Omega
\end{aligned}
\end{equation}
with periodic boundary conditions. For inverse evolution problems, one can also employ different numerical methods to solve them and determine the values of numerical solutions at different time point. For example, if solving the Eq. \eqref{inverse_pde} by the Forward Euler method for temporal discretization and a finite difference scheme for spatial discretization, we can obtain the following discrete numerical scheme for inverse evolution:
\begin{equation}
\label{scheme_inverse}
    \frac{U^{n+1}-U^n}{\Delta t} = -F(U^n)
\end{equation}
Here $U^n$ and $U^{n+1}$ respectively denote the numerical solution at time $T_{n}$ and $T_{n+1}$, where $T_{n+1}-T_n = \Delta t$. $F$ represents finite difference approximations of $\mathcal{F}$, and $F(U^n)$ can be conceptualized as a combination of convolutions of $u$ with designated filters.

\paragraph{Data Generation Schemes}In this part, we demonstrate how to use inverse evolution to produce data consistent with an implicit method without solving any implicit problems.

Let us consider the Forward Euler method for the inverse evolution, i.e. scheme \eqref{scheme_inverse}. Given $U^n$, if we compute with a large time step $\Delta t$, we can efficiently obtain $U^{n+1}$, although $(U^n, U^{n+1})$ may not accurately satisfy the inverse evolution equation. Introducing a new pair $(V^{n}, V^{n+1})$ by interchanging the order of $U^n$ and $U^{n+1}$:
\begin{equation}
\label{order_change}
    (V^{n}, V^{n+1}):= (U^{n+1}, U^n),
\end{equation}
then $(V^{n}, V^{n+1})$ can be interpreted as pairs originating from the original forward equation. 

By substituting Eq. \eqref{order_change} into Eq. \eqref{scheme_inverse}, we derive
\begin{equation}
  \frac{V^{n+1}-V^{n}}{\Delta t} = F(V^{n+1}).
\end{equation}
This formula implies that the pair $(V^{n}, V^{n+1})$, i.e. $(U^{n+1}, U^n)$, satisfies the formula of the implicit Backward Euler method derived from the original equation. Given $U^{n}$, we can directly compute $U^{n+1}$ according to \eqref{scheme_inverse} without solving any nonlinear equations. Subsequently, we can utilize $(U^{n+1}, U^n)$, which perfectly conforms to the implicit Euler formula, as training data for the neural operator.

The data pair $(U^{n}, U^{n+1})$ results from the Forward Euler method tailored to the inverse evolution, while its reverse pair $(U^{n+1}, U^n)$ precisely satisfies the Implicit Euler method of the original evolution. This ensures the reliability of the acquired data for training. With sufficient $U^{n}$, we can efficiently obtain enough data pairs $(U^{n+1}, U^{n})'s$. It is noteworthy that the favorable attributes of implicit methods provide us with the flexibility to choose a larger time step, $\Delta t$, rather than being confined to $O(h^2)$. This flexibility significantly enhances computational efficiency while maintaining accuracy of the generated data.
\paragraph{Initialization for Inverse Evolution}
While this approach appears promising, there remains a significant challenge that needs to be addressed: obtaining an appropriate initial value, i.e. $U^{n+1}$, for the inverse evolution process. One option is to choose random initial values for $U^{n+1}$. However, it is important to note that the inverse evolution is inherently unstable. This instability means that a random initialization is likely to result in $U^{n+1}$ transitioning to a more irregular, or even singular state $U^{n}$. Consequently, the obtained data pairs may lack practical physical significance and may not adequately capture the characteristics of a desirable solution space.

To address the initialization challenge, we propose a method where we randomly combine the values of solutions at different time points and utilize them as the initialization for the inverse evolution. Specifically, let $\mathbf{U}:=(U^1,U^2,\cdots,U^n)$ represent a time series from the original data. And let $\{\mathbf{U}_i\}_{i=1}^n$ denote the set of $n$ different time series. By performing different random sorts on the samples in $\{\mathbf{U}_i\}_{i=1}^n$ for $k$ times, we obtain $\{\mathbf{U}_{R_1(i)}\}_{i=1}^n,\{\mathbf{U}_{R_2(i)}\}_{i=1}^n,\cdots,\{\mathbf{U}_{R_k(i)}\}_{i=1}^n$. Here $R_k(\cdot)$ is the permutation mapping corresponding to the $k$-th random sort. Then, we set $\{\mathbf{U}_i^*\}_{i=1}^n$ as the initialization such that 
\begin{equation}
\label{initialization}
    \mathbf{U}_i^{*} = \sum_{j=0}^k \lambda_j \cdot\mathbf{U}_{R_j(i)} + C,
\end{equation}
where $\lambda_j$ and $C$ are constants, and $\mathbf{U}_{R_0(i)}: = \mathbf{U}_i$.
We anticipate that the $\{\mathbf{U}_i^*\}_{i=1}^n$ will span the solutions at various time points and differ from the original data $\{\mathbf{U_i}\}_{i=1}^n$. By using elements of $\mathbf{U}_i^{*}$ at different time as initializations, we can compute inverse evolution to generate training data pairs of different time. Importantly, although the initializations are derived from combinations of $\{\mathbf{U_i}\}_i^n$, the resulting data pairs are distinct due to the nonlinear nature of both the original and inverse evolution processes. This nonlinearity ensures that the solutions obtained after one time step from a linear space will differ significantly from the original space. Therefore, the generated data pairs do not overlap with the original data. With a small set of original data points, we can apply different $\lambda_j$ and $C$ values to create a large number of initializations. These initializations can then be used to generate a diverse and extensive set of training pairs for the FNO. 

\begin{figure}
    \centering
    \includegraphics[width=0.25\linewidth]{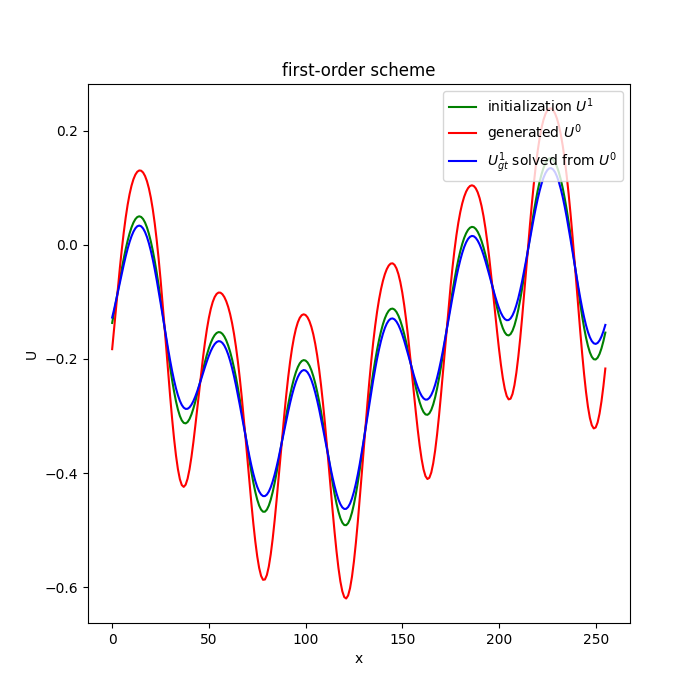}
    \includegraphics[width=0.25\linewidth]{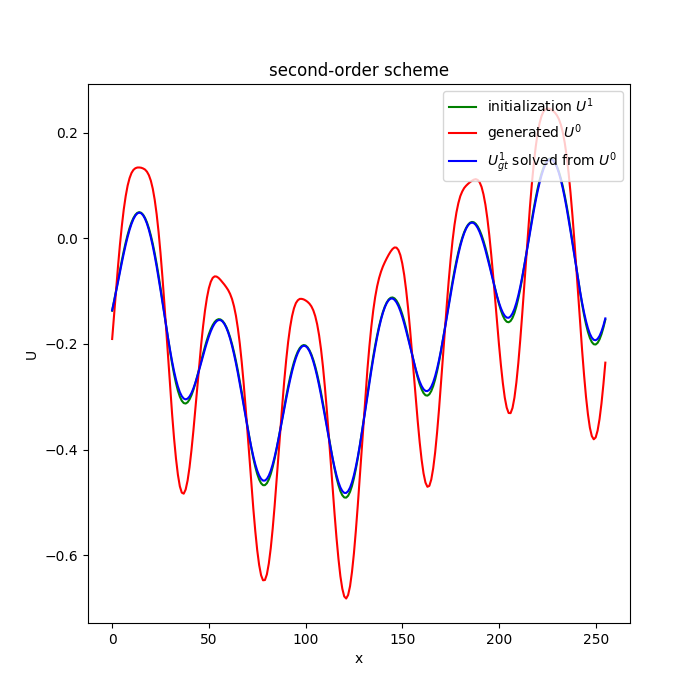}
    \includegraphics[width=0.25\linewidth]{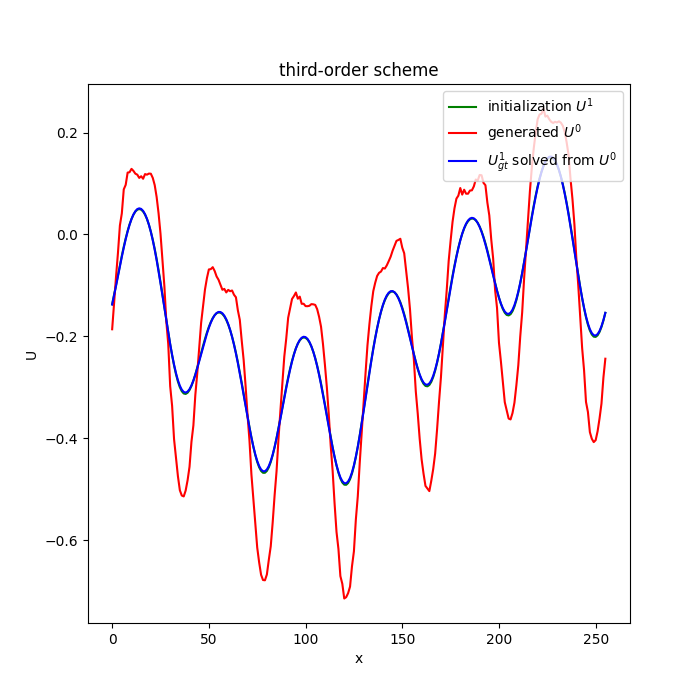}\\
    \includegraphics[width=0.25\linewidth]{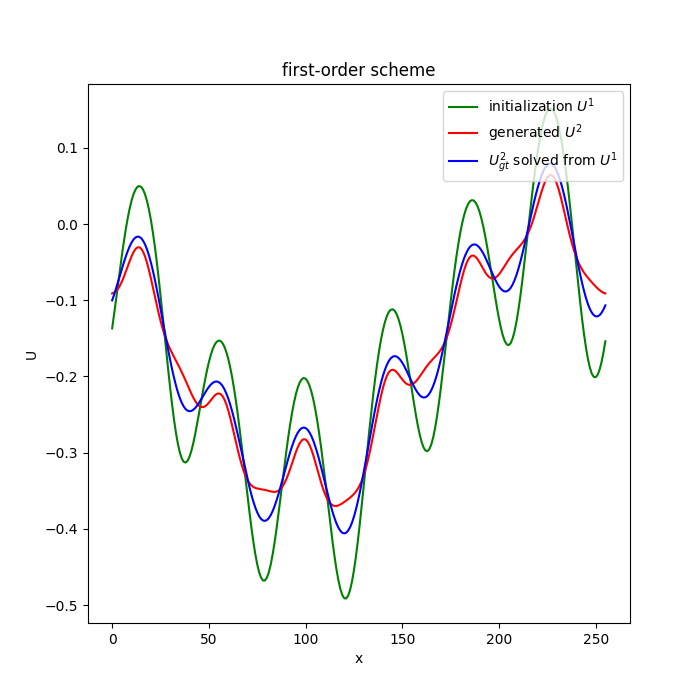}
    \includegraphics[width=0.25\linewidth]{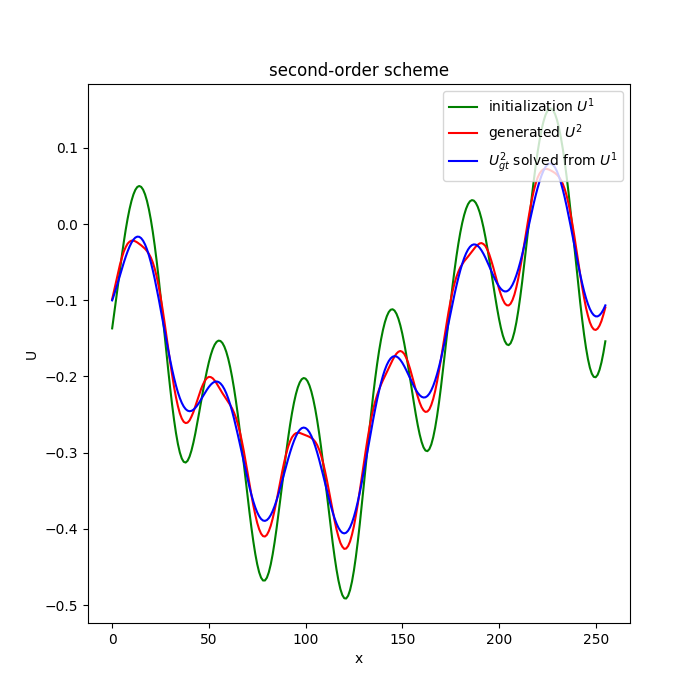}
    \includegraphics[width=0.25\linewidth]{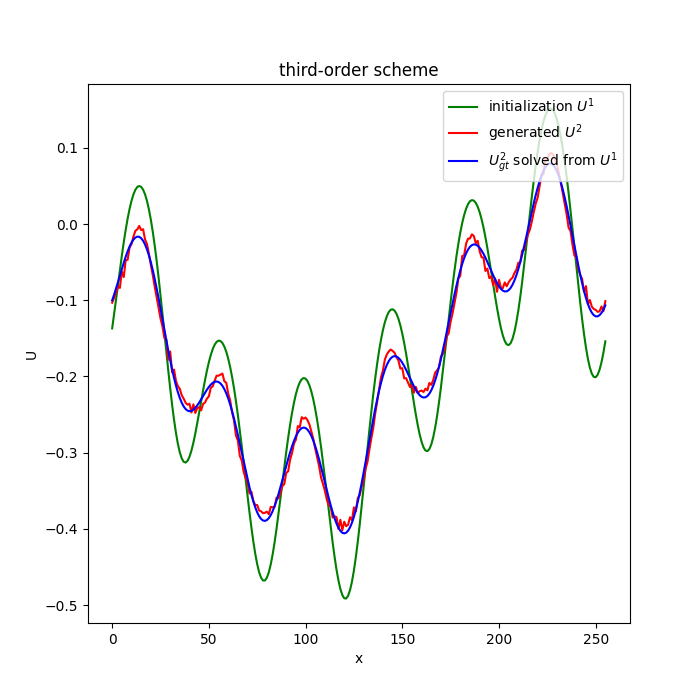}
    \caption{Comparison between inverse evolution and forward evolution on heat diffusion equation with $\Delta t =0.05$. Inverse evolution produces more accurate data pairs ($U_0, U_1$) as the order increases, while forward evolution yields inaccurate data pairs ($U_1, U_2$) across all schemes.}
    \label{fig:ie_vs_fe}
\end{figure}

\paragraph{High-order Schemes for Inverse Evolution}
In practical implementation, we have found that the data generated through the Forward Euler method-based inverse evolution are not sufficiently accurate, as shown in the top left corner of Figure \ref{fig:ie_vs_fe}. This inaccuracy arises because the Forward Euler method is a first-order numerical scheme. To obtain more accurate data, we have explored higher-order methods for the inverse evolution given as follows:
\paragraph{First-order scheme}
\begin{equation}
   U^{n+1} =  U^{n} - \Delta t U_t^{n},
\end{equation}

\paragraph{Second-order scheme}
\begin{equation}
   U^{n+1} =  U^{n} - \Delta t U_t^{n} + \frac{\Delta t^2}{2} U_{tt}^{n},
\end{equation}

\paragraph{Third-order scheme}
\begin{equation}
	U^{n+1} =   U^{n} - \Delta t U_t^{n} + \frac{\Delta t^2}{2} U_{tt}^{n} - \frac{\Delta t^3}{6} U_{ttt}^{n},
 \end{equation}

where
\begin{align}
    &U_t = F(U),\\
    &U_{tt} = F'(U)U_t,\\
    &U_{ttt} = F'(U)U_{tt} + F''(U)U_t^2.
\end{align}
Similarly, it can be proved that the data pairs acquired from the high-order scheme derived from the Taylor expansion also satisfy the corresponding high-order implicit schemes. The comparison between different schemes in terms of the accuracy is displayed in Table \ref{table_acc_test}. Notably, one can also derive schemes with higher order through the Taylor expansion. Given a sufficiently small $\Delta t$, higher-order schemes yield more accurate data. However, higher-order schemes also tend to be more unstable due to the instability arising from the computation of high-order derivatives of $U$. Therefore, when handling very fine meshes, we recommend using relatively small time steps and second-order schemes. For coarser meshes, third-order or higher schemes with relatively larger time steps are advisable.

\paragraph{Inverse Evolution vs. Original Evolution} As discussed previously, the data pairs obtained via inverse evolution come from implicit numerical methods derived from the original evolution, making them suitable as training data. Figure \ref{fig:ie_vs_fe} and Table \ref{table_acc_test} can further demonstrate the effectiveness of data generated through inverse evolution. However, employing similar explicit schemes on the forward evolution to generate data is not suitable. While both inverse and forward evolutions exhibit instability, the nature of this instability differs. Inverse evolution instability affects the input data, whereas forward evolution instability impacts the output data. Consequently, data pairs from inverse evolution can train neural operators to map unstable input to stable output, whereas data pairs from forward evolution would compel neural operators to map stable input to unstable output. This distinction makes the two approaches fundamentally opposite. Additionally, even though the generated input data might be irregular and contain some noise, it can still evolve into the generated output data according to the given PDE since the data pairs conform to implicit numerical schemes of the PDE. For forward evolution, the instability of explicit schemes makes it nearly impossible to obtain accurate data pairs with a large $\Delta t$. The difference is illustrated in Figure \ref{fig:ie_vs_fe}. 

\paragraph{Preprocessing for Data with Sharp Interfaces}We observe that solutions with poor differentiability can exacerbate the instability of the inverse evolution. For instance, solutions of Burgers' equation often exhibit sharp interfaces, particularly when the viscosity approaches zero. During such instances, the solution may contain abrupt, nearly discontinuous changes, leading to poor properties for a numerical solution, such as very high gradients. As a consequence, the inverse evolution may struggle to provide accurate training pairs. One approach to address this is by reducing the spatial mesh size $h$, and hence $\Delta t$ to ensure stability. However, in many cases, there is a preference for a relatively large time interval between generated data pairs. An alternative way is to smooth the initialization. The inherent instability can be somewhat mitigated by smoothing techniques. However, while smoothing can alleviate the instability caused by sharp interfaces, it also tends to smooth out the sharp interfaces of the solutions. Consequently, the generated data may not effectively assist neural operators in learning to predict solutions with sharp interfaces.

\begin{figure}
    \centering
    \includegraphics[width=0.25\linewidth]{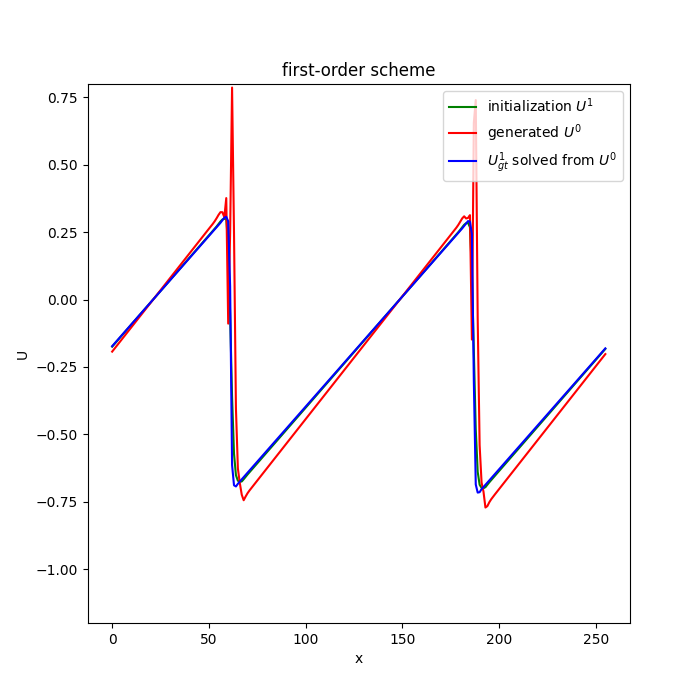}
    \includegraphics[width=0.25\linewidth]{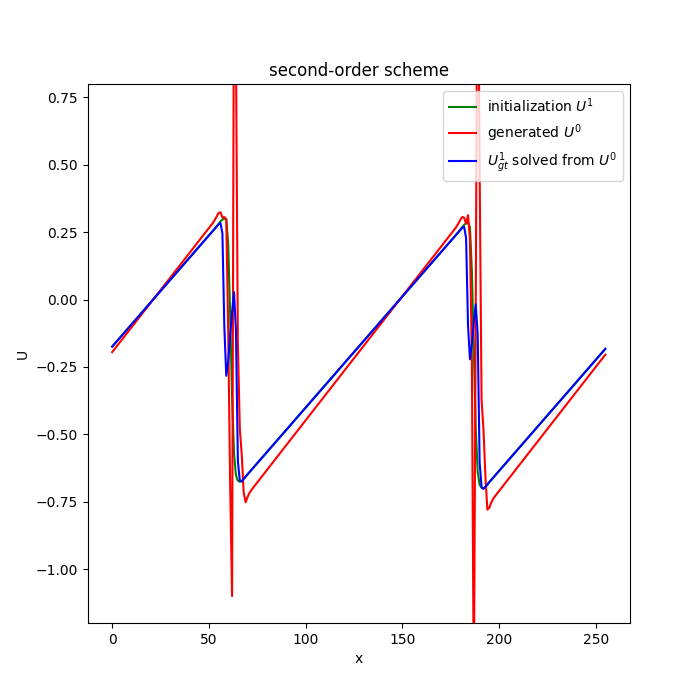}
    \includegraphics[width=0.25\linewidth]{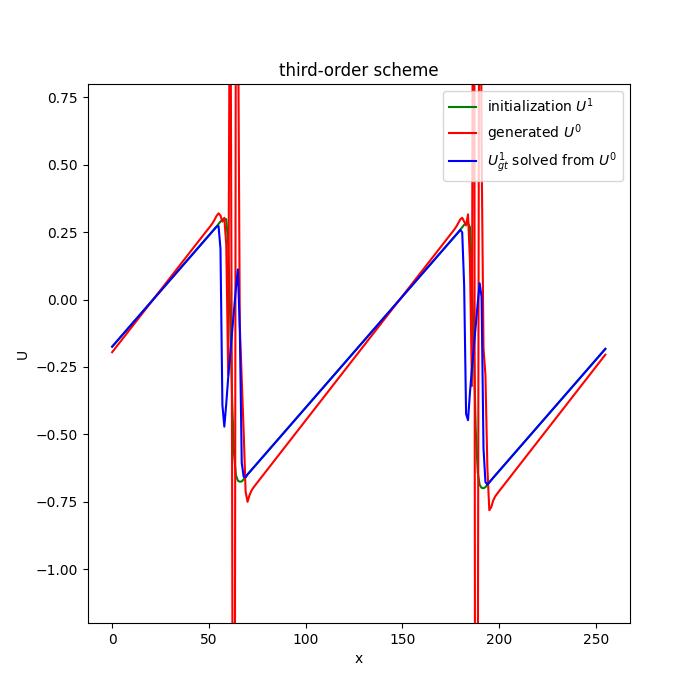}\\
    \includegraphics[width=0.25\linewidth]{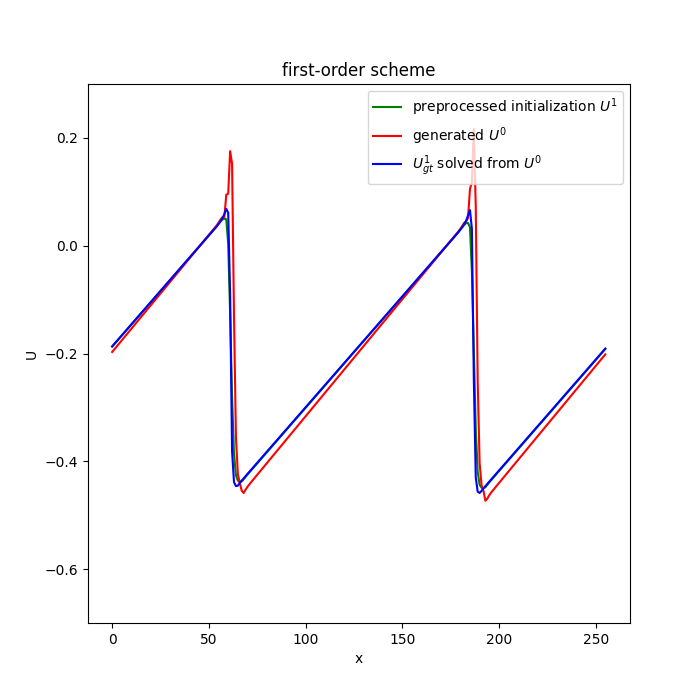}
    \includegraphics[width=0.25\linewidth]{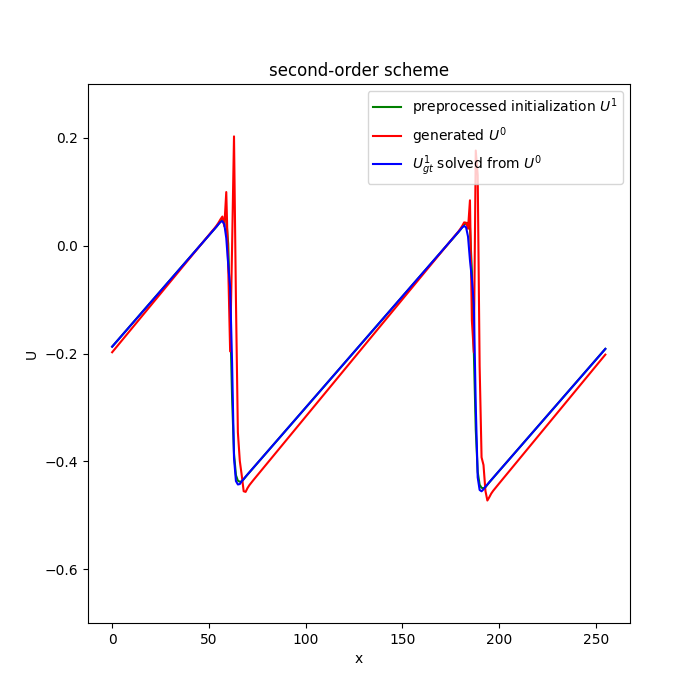}
    \includegraphics[width=0.25\linewidth]{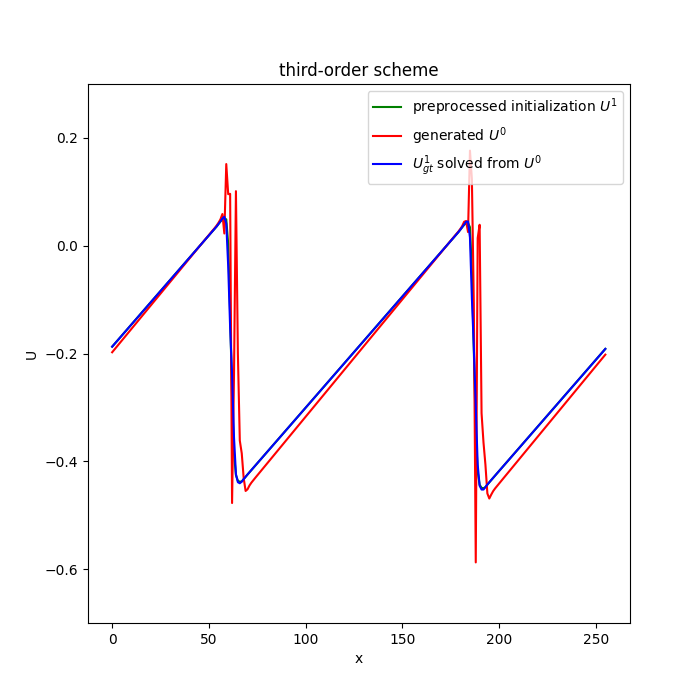}
    \caption{Inverse evolution between original data and preprocessed data. The first row shows the inverse evolution of the original data, while the second row presents the results based on preprocessed data.}
    \label{fig:preprocess}
\end{figure}

To overcome this challenge, we propose normalization and rescaling techniques to preprocess our generated initializations. We find that normalization and rescaling can alleviate instability issues without reducing $\Delta t$ or resorting to smoothing, which can be observed in Figure \ref{fig:preprocess} It is worth noting that normalization and rescaling techniques can still preserve the sharp interface, which makes the preprocessed data more valuable than smoothed data. Given an initial value $U$ for the inverse evolution, the preprocessed $U$ can be expressed as:
\begin{equation}
\label{eq:preprocess}
    U^{pre} = a\cdot (U - \text{mean}(U)) + C,
\end{equation}
where $a\in [0, 1]$ used for rescaling and $C$ is a constant, and $\text{mean}(U)$ is the mean value of $U$. 

\modified{It is important to note that normalization and rescaling preprocessing are only applied to initializations when the solution requires consideration of very fine spatial structures, such as in the case of the Burgers’ equation with a very small diffusion coefficient. In addition, the rescaling in the preprocessing method is a form or r-adaptivity in which the local mesh points are rescaled to give a more accurate local solution. It is known (see the book \cite{huang2010adaptive}) that this is effective in reducing solution error.}

\paragraph{Spatial Discretization for Inverse Evolution} In addition to the temporal discretization, the spatial discretization is also crucial in the numerical scheme. Various numerical methods have
been developed for spatial discretization, including finite difference, finite element, finite volume and spectral methods (see \cite{quarteroni2008numerical}). For inverse evolution, we employ finite difference methods due to their ease of implementation on regular domains. Additionally, we test the pseudo-spectral method, and both methods exhibit comparable performance. However, when testing the Navier-Stokes equation, we observe that the accuracy of data pairs generated using the finite difference method is sensitive to the constant in Eq. \eqref{initialization}, whereas the pseudo-spectral method is not. More precisely, as the absolute value of the constant changes, the stability and accuracy of the data pairs generated using the finite difference method changes apparently. Conversely, for the pseudo-spectral method, changes of the constant have minimal impact on data stability and accuracy. This suggests that when using the pseudo-spectral method, we have flexibility to choose different and relatively large constants for data augmentation. This flexibility becomes particularly valuable when preprocessing is required, as preprocessed data often have smaller scales compared to the original data. By employing a large constant, we can expand the range of generated data, enabling them to encompass a broader space. Therefore, when dealing with the Navier-Stokes equation, we adopt the pseudo-spectral method for spatial discretization to ensure greater stability and higher accuracy for the data augmentation.

\section{Experiments}
\label{experiments}
\subsection{Accuracy Test}
\begin{figure}
    \centering
    \includegraphics[width=0.2\linewidth]{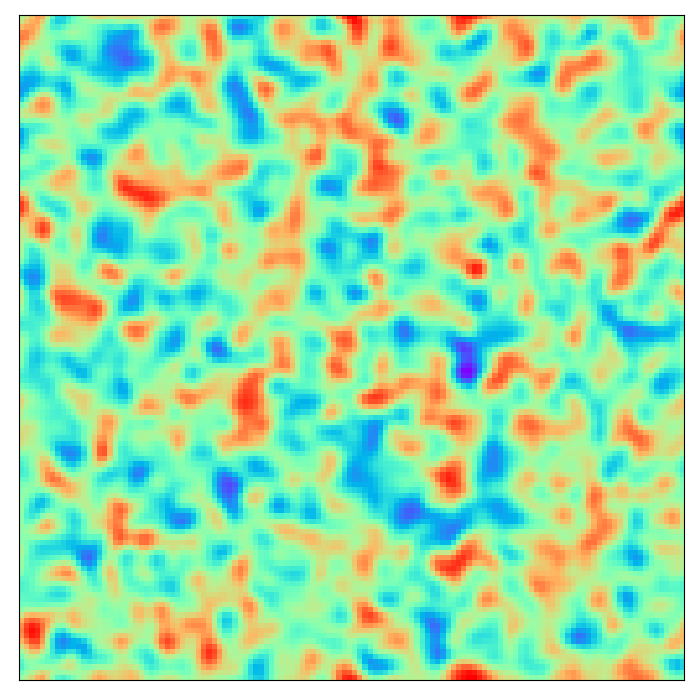}
    \includegraphics[width=0.2\linewidth]{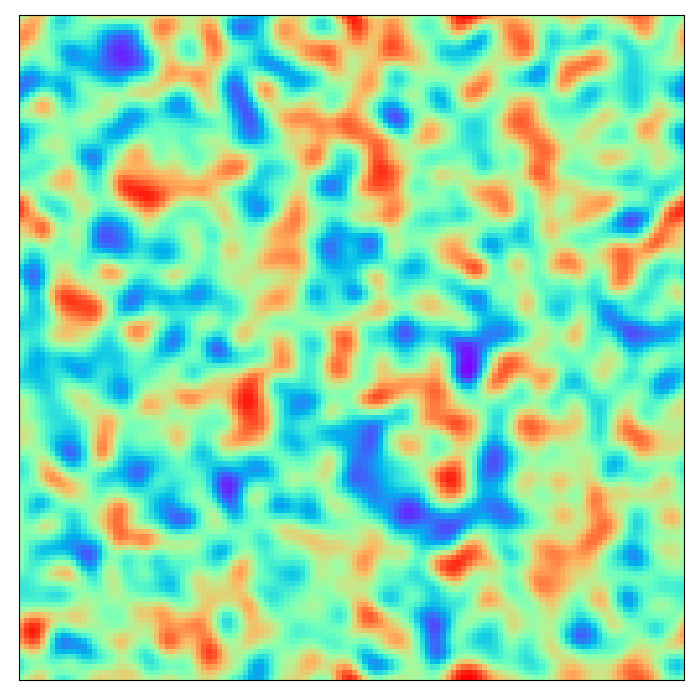}
    \includegraphics[width=0.2\linewidth]{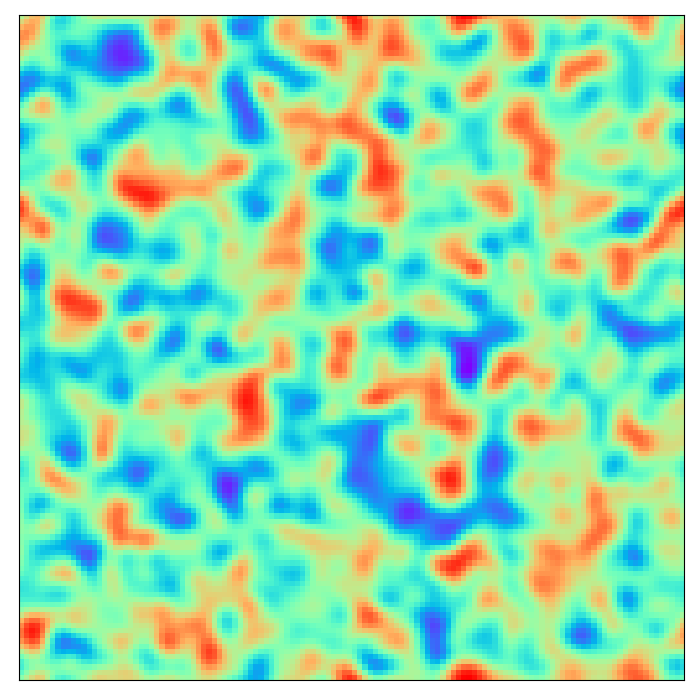}\\\vspace{0.5mm}\hspace{0.1mm}
    \includegraphics[width=0.2\linewidth]{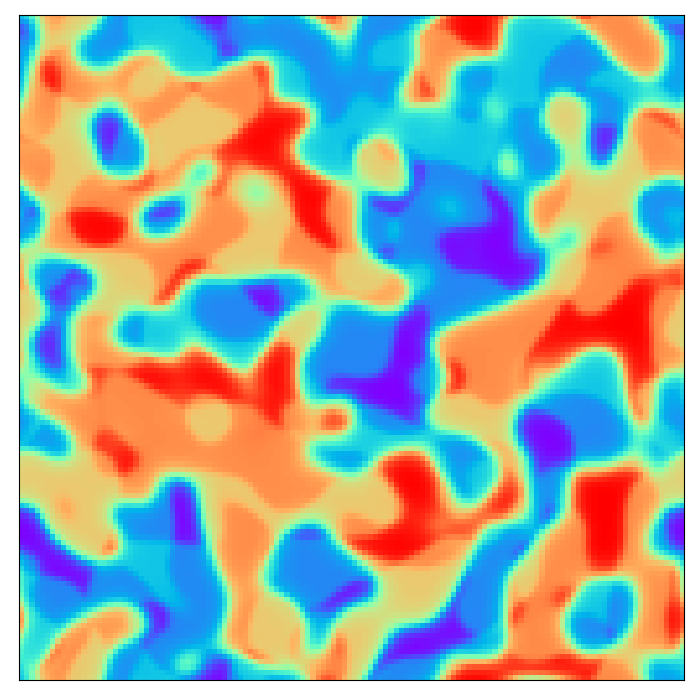}
    \includegraphics[width=0.2\linewidth]{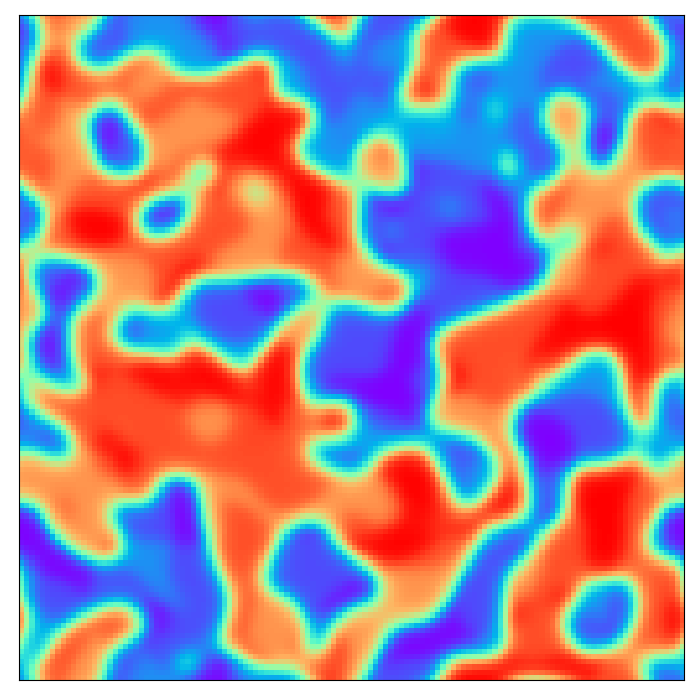}
    \includegraphics[width=0.2\linewidth]{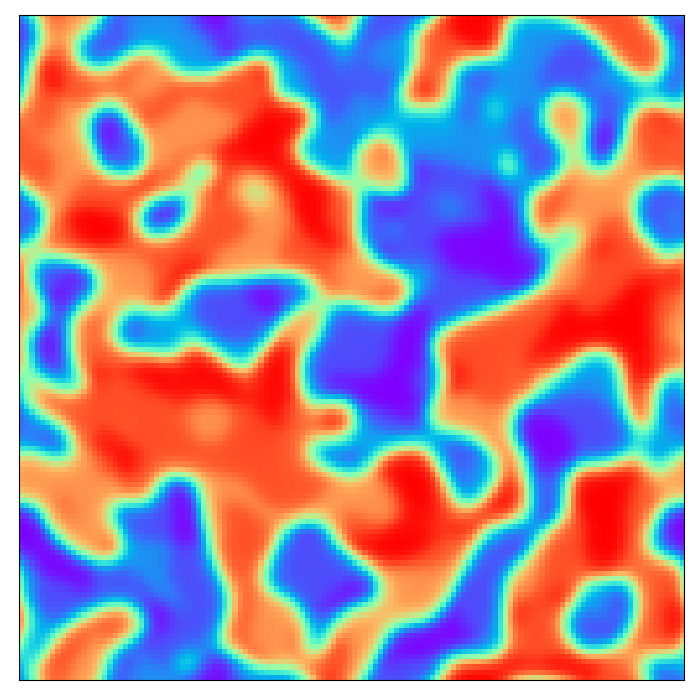}
    \caption{Generated data for Allen-Cahn equation with $\epsilon=0.01$ and $\Delta t=0.5$. The first two columns are generated inputs and outputs from the inverse evolution, respectively. The last column shows the true solutions of the generated input.}
    \label{fig:allen-cahn_0.01}
\end{figure}

\begin{figure}
    \centering
    \includegraphics[width=0.2\linewidth]{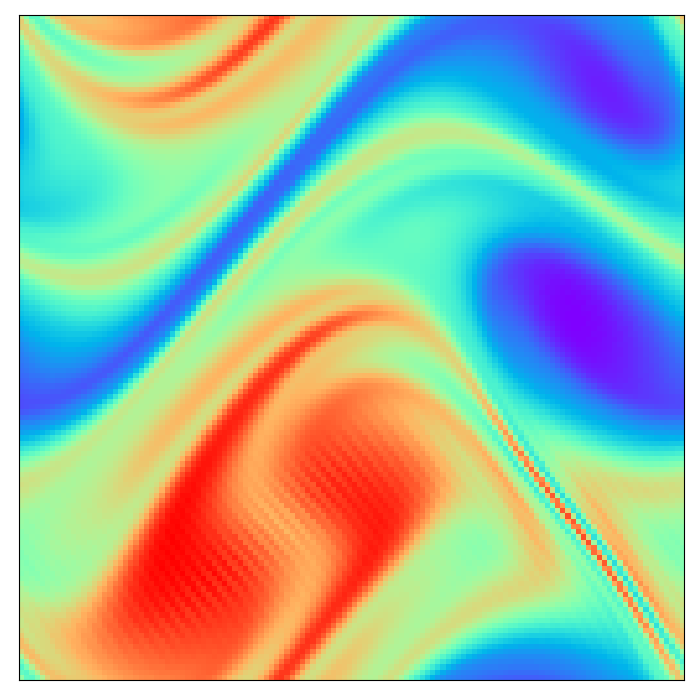}
    \includegraphics[width=0.2\linewidth]{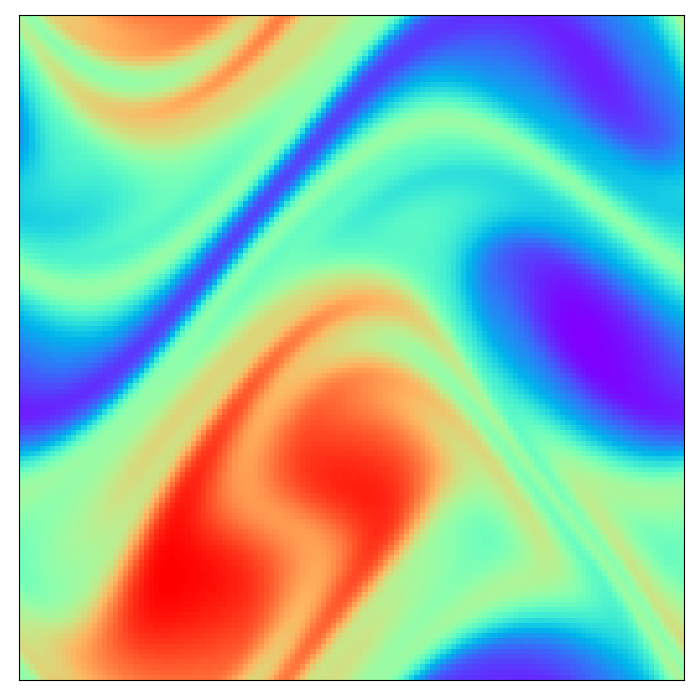}
    \includegraphics[width=0.2\linewidth]{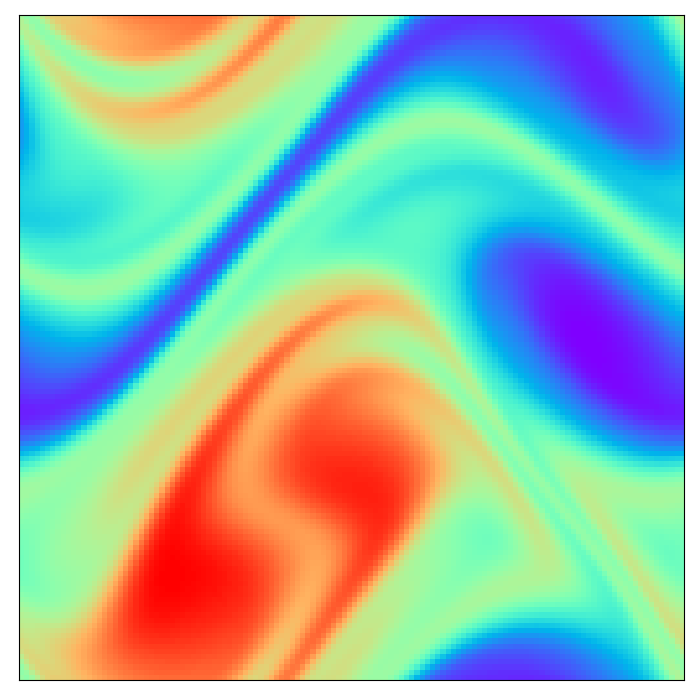}\\\vspace{0.5mm}\hspace{0.1mm}
    \includegraphics[width=0.2\linewidth]{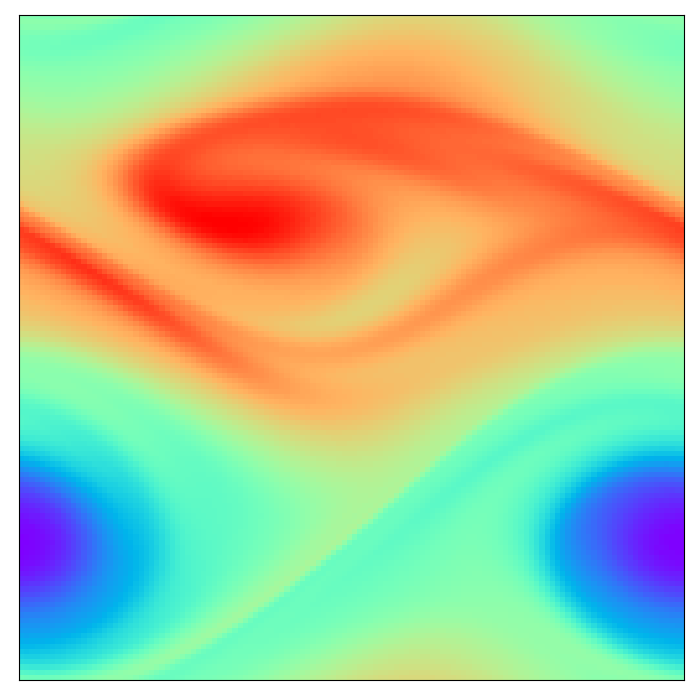}
    \includegraphics[width=0.2\linewidth]{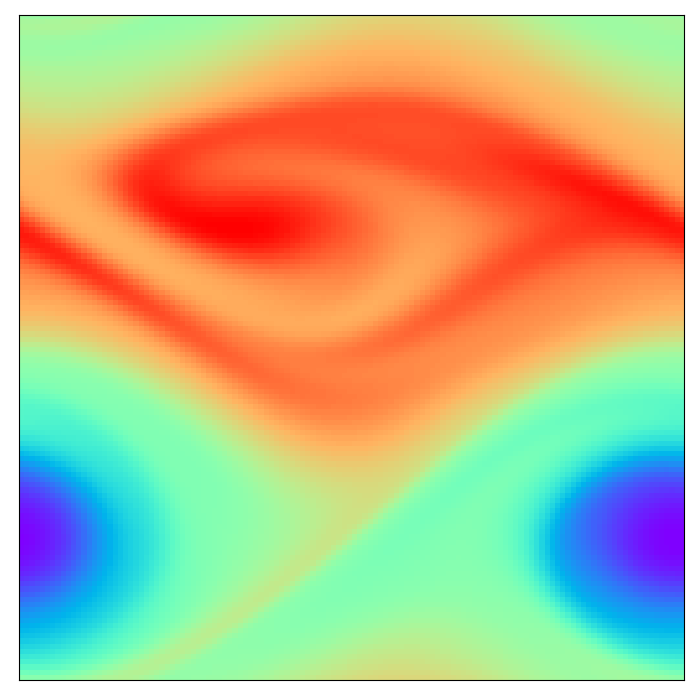}
    \includegraphics[width=0.2\linewidth]{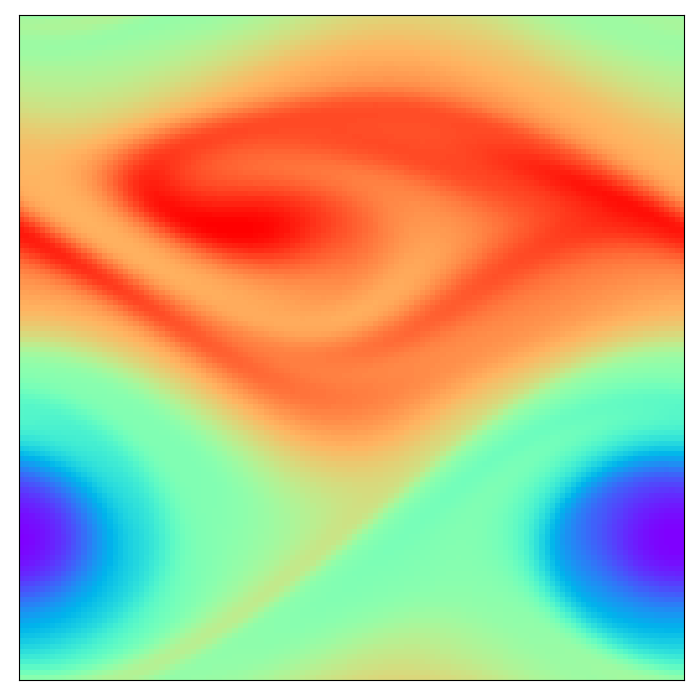}
    \caption{Generated data for Navier-Stokes equation with $\nu = 0.001$ and $\Delta t=0.5$. The first two columns are generated inputs and outputs from the inverse evolution, respectively. The last column shows the true solutions of the generated input.}
    \label{fig:navier stokes_0.0001}
\end{figure}

In this part, we evaluate the accuracy of our data augmentation method across four equations with periodic boundary conditions: the heat diffusion equation, Burgers' equation, the Allen-Cahn equation, and the Navier-Stokes equation. The heat diffusion in 1D is given by $u_t = u_{xx}$, and the specific formulas for other equations are given in Eq. \eqref{eq:burgers}, Eq. \eqref{eq:ac} and Eq. \eqref{eq:ns}. The accuracy is assessed under varying data resolutions, time intervals ($\Delta t$), and inverse evolution schemes. The results of test accuracy are displayed in Table \ref{table_acc_test}. We also provide some samples of generated data in Figure \ref{fig:allen-cahn_0.01} and Figure \ref{fig:navier stokes_0.0001}. Our results indicate that the generated data exhibit high accuracy, particularly when $\Delta t$ is not large. Furthermore, accuracy tends to increase with the order of the schemes. Changes in data resolution have minimal impact on accuracy; however, increasing resolution can exacerbate the instability of high-order schemes, especially when $\Delta t$ is large. This instability can significantly affect accuracy. Therefore, for high-resolution data, we recommend using relatively small $\Delta t$ values with second-order schemes to balance the time interval, accuracy, and stability.

\begin{table}
\caption{Accuracy test for data generated by inverse evolution. The error is evaluated by the relative error on $L^2$ norm. We put — when error fails to reach below 1.0. }
\label{table_acc_test}
\centering
\resizebox{0.8\textwidth}{!}{
\begin{tabular}{p{0.8cm}<{\centering}|c|p{1cm}<{\centering}|p{1cm}<{\centering}|p{1.5cm}<{\centering}p{1.5cm}<{\centering}p{1.5cm}<{\centering}l}
\toprule
Dim & PDEs & $\Delta t$ &Res. & $1^{st}$-order & $2^{nd}$-order & $3^{rd}$-order& Magnitude\\
\bottomrule
\multirow{12}{*}{1D} & \multirow{6}{*}{Heat Diffusion}& \multirow{3}{*}{0.01} &256 & $3.918$ & $1.032$ & \textbf{0.904}  & \multirow{6}{*}{$\times 10^{-4}$}\\
& & & 512& $3.967$ & $0.881$ & \textbf{0.797}  &\\
& & & 1024&$4.004$ & $0.796$ & \textbf{0.715}  &\\
\cmidrule(lr){3-7}
& & \multirow{3}{*}{0.05} & 256 & $75.81$ & $12.80$ & \textbf{5.709}  &\\
& & & 512& $76.45$ & $12.74$ & \textbf{5.049}  &\\
& & & 1024& $78.94$ & $10.32$ & \textbf{2.729}  &\\
\cmidrule(lr){2-8}
& \multirow{6}{*}{\makecell{Burgers \\($\nu=0.1$)}} & \multirow{3}{*}{0.01} & 256 & $44.48$ & $4.651$ & \textbf{1.247} & \multirow{6}{*}{$\times 10^{-5}$} \\
& & & 512& $26.26$ & $0.810$ & \textbf{0.731}  &\\
& & & 1024& $31.47$ & \textbf{0.813} & $185.2$  &\\
\cmidrule(lr){3-7}
& & \multirow{3}{*}{0.05} & 256 & $58.13$ & \textbf{6.039} & 6.221  &\\
& & & 512& $58.99$ & \textbf{5.713} & $657.5$  &\\
& & & 1024& $61.61$ & \textbf{7.305} & $-$  &\\
\cmidrule(lr){1-8}
\multirow{12}{*}{2D} & \multirow{6}{*}{\makecell{Allen-Cahn\\ ($\epsilon=0.05$)}} & \multirow{3}{*}{0.1} & $128$ & $16.23$ & $0.996$ & \textbf{0.132}  & \multirow{6}{*}{$\times 10^{-4}$}\\
& & & $256$ & $17.34$ & $0.961$ & \textbf{0.143}  &\\
& & & $512$ & $18.78$ & $0.938$ & \textbf{0.474}  &\\
\cmidrule(lr){3-7}
& & \multirow{3}{*}{0.5} & $128$ & $393.1$ & $88.37$ & \textbf{20.06} &\\
& & & $256$& $455.1$ & $92.01$ & \textbf{20.41}  &\\
& & & $512$& $524.1$ & \textbf{98.21} & $-$  &\\
\cmidrule(lr){2-8}
& \multirow{6}{*}{\makecell{Navier-Stokes\\ ($\nu=0.001$)}}  & \multirow{3}{*}{0.1} & $128$ & $3.332$ & 1.621 & \textbf{1.505} & \multirow{6}{*}{$\times 10^{-4}$}\\
& & & $256$& $2.968$ & 0.346 & \textbf{0.389} &\\
& & & $512$& $3.015$ & 0.432 & \textbf{0.410}  &\\
\cmidrule(lr){3-7}
& & \multirow{3}{*}{0.5} & $128$ & $65.71$ & $14.32$ & \textbf{6.240} &\\
& & & $256$& $63.87$ & $13.51$ & \textbf{5.081}  &\\
& & & $512$& $63.86$ & $13.73$ & \textbf{9.853}  &\\
\bottomrule
\end{tabular}
}
\end{table}

\subsection{Computational Efficiency Analysis}
In this section, we assess the computational efficiency of our data augmentation process by comparing the runtime of our method against the traditional forward Euler method for generating 100 data pairs. The runtime comparison is presented in Table \ref{table_forward_vs_ie}. It is important to emphasize that the measured times are for 100 data pairs, while typical datasets for training and testing consist of tens of thousands of pairs. As a result, the difference in computational efficiency would be significantly magnified in larger datasets. Moreover, higher-resolution and higher-dimensional data would further exacerbate the disparity in computational demands.
\begin{table}
    \centering
    \caption{Time required (in seconds) to generate 100 data pairs at a resolution of 256. Our method significantly outperforms the traditional explicit forward solver. This difference is expected to increase with higher resolutions and dimensions.}
    \label{table_forward_vs_ie}
    \resizebox{0.8\textwidth}{!}{\begin{tabular}{c|c|c|c}
    \toprule
    Methods & Burgers' Equation (1D) & Allen-Cahn Equation (2D) & Navier-Stokes Equation (2D) \\ \midrule
    Explicit Forward Solver & 3.218 & 11.800 & 74.831 \\ \midrule
    Ours & \textbf{0.2727} & \textbf{0.1541} & \textbf{0.0325} \\  
    \bottomrule
    \end{tabular}}
    \end{table}

\subsection{FNO with Inverse Evolution Data Augmentation}
In this part, we assess the effectiveness of our data augmentation technique on the Fourier Neural Operator (FNO), a state-of-the-art neural operator designed for efficiently solving partial differential equations (PDEs). FNO utilizes Fourier transforms to project the input function into the frequency domain, enabling the model to capture global information more effectively. In the frequency domain, learnable filters are applied to modify specific frequency components, allowing the model to adjust its representations during training. The FNO architecture has demonstrated superiority in solving a variety of PDEs, achieving high accuracy and computational efficiency.

We conduct tests on three equations: Burgers' equation, the Allen-Cahn equation, and the Navier-Stokes equation with periodic boundary conditions. The FNO is implemented using the code available at \href{https://github.com/neuraloperator/neuraloperator}{https://github.com/neuraloperator/neuraloperator}. For all experiments, FNOs are trained for 500 epochs with the Adam optimizer. The learning rate is set to 0.001 for Burgers' equation and 0.0025 for the Allen-Cahn and Navier-Stokes equations. We apply a learning rate weight decay of 1e-4 across all experiments, with the learning rate decaying by a factor of 0.5 after every 100 epochs. The number of Fourier modes is set to 32 for Burgers' equation and 48 for Navier-Stokes equation. For the Allen-Cahn equation, the modes are set to 32 when \(\epsilon=0.05\) and 64 when \(\epsilon=0.01\). The numbers of hidden and projection channels are (24, 24) for Burgers' equation, (64, 128) for the Allen-Cahn equation, and (64, 256) for the Navier-Stokes equation. The batch size for all experiments is 64. All experiments are conducted on a single NVIDIA A100 GPU. For all experiments, we employ the third-order inverse evolution schemes for data augmentation. The results are presented in Table \ref{table_fno_iedata}. 

\paragraph{Burgers’ Equation} Burgers' equation is a nonlinear PDE with various applications including modeling shock formation in fluid dynamics. In 1D it is given by the following PDE: 
\begin{equation}
\label{eq:burgers}
\begin{aligned} 
    u_t  &= -u u_x + \nu/\pi \Delta u, &&\quad x\in[0,1], t\in(0,T), \quad 0 < \nu \ll 1,\\
    u(0,x) &= u_0(x), &&\quad x\in[0,1]
\end{aligned} 
\end{equation}
where $u_0\in L^2_{per}((0,1), \mathbb{R})$ is the initial state and $\nu\in \mathbb{R}_+$ is the diffusion coefficient. In our experiments, $\nu$ is chosen to be 0.1 and 0.001 and the final time $T=2$. We aim to learn the operator mapping $u(t, \cdot)\to u(t + 0.05, \cdot)$. We solve Burgers' equation using the same method as \cite{takamoto2022pdebench} to generate original data, and the code can be found in \href{https://github.com/pdebench/PDEBench}{https://github.com/pdebench/PDEBench}. 1000 time series of resolution $(41, 256)$ are solved, with 500 allocated to the training set and 500 to the test set, respectively. Preprocessing is applied to the initialization of inverse evolution augmentation when $\nu=0.001$. The rescaling coefficient is 1 and $C$ is randomly chosen between $[-0.1, 0.1]$ for each initial value.

\paragraph{The Allen-Cahn Equation}
The 2D Allen-Cahn equation is a nonlinear PDE that can be used to describe the phase separation process. It takes the form
\begin{equation}
\label{eq:ac}
\begin{aligned}
    u_t &= u - u^3 + \epsilon^2\Delta u, && \mathbf{x}\in[0,1]^2, t\in(0,T)\\
    u(0,\mathbf{x}) &= u_0(\mathbf{x}), && \mathbf{x}\in[0,1]^2
\end{aligned}
\end{equation}
where $u_0\in L^2_{per}([0,1]^2, \mathbb{R})$ is the initial value and $0 <\epsilon \ll 1$ is a constant related to the thickness of the interface between different phases. In our experiments, $\epsilon$ is chosen to be 0.01 and 0.05. And the operator mapping we learn is that from $u(0, \cdot)\to u(0.5, \cdot)$ and the final time $T=16$. We generate the original data using Forward Euler with time-step of 1e-4. 800 time series of resolution $(33, 128, 128)$ are solved, with 400 used for the training set and 400 for the test set, respectively.

\paragraph{The Navier-Stokes Equation} We consider the vorticity-stream form of Navier-Stokes equation for a 2D incompressible fluid:
\begin{equation}
\label{eq:ns}
    \begin{aligned}
    \omega_t &= - u\cdot\omega_x - v\cdot\omega_y + \nu\Delta\omega + f, && \mathbf{x}\in[0,1]^2, t\in(0,T)\\
    \omega &= v_x - u_y, \quad w(0,\mathbf{x}) = w_0(\mathbf{x}), && \mathbf{x}\in[0,1]^2,
\end{aligned}
\end{equation}
where $w_0\in L^2_{per}([0,1]^2, \mathbb{R})$ is the initial value, $0< \nu \ll 1$ is the viscosity coefficient, $(u,v)$ represents the velocity and $f$ is a force term. In our experiment, we test the cases $\nu = 0.001$ and $\nu =0.0001$ with a fixed force term $f = 0.1(\sin(2\pi(x + y)) + \cos(2\pi(x + y)))$ and the final time $T=50$. We learn the operator mapping from $\omega(0, \cdot)\to \omega(0.5, \cdot)$. The original data are generated based on pseudo-spectral method and Crank–Nicolson method with a time-step of 1e-4. We solved 400 time series of resolution $(100, 128, 128)$, with 200 for the training set and 200 for the test set, respectively. Pseudo-spectral method is employed for the spatial discretization for the inverse evolution. 
\begin{table}
\centering
\caption{Test relative error on $L^2$ norm for FNO trained on different datasets.}
\label{table_fno_iedata}
\resizebox{0.8\textwidth}{!}{\begin{tabular}{c|c|llllll}
\toprule
 \multirow{2}{*}{Data Size}& \multirow{2}{*}{Augment.}& \multicolumn{2}{l}{Burgers' Equation} & \multicolumn{2}{l}{Allen-Cahn Equation} & \multicolumn{2}{l}{Navier-Stokes Equation} \\\cmidrule(lr){3-4}\cmidrule(lr){5-6}\cmidrule(lr){7-8}
 & & $\nu = 0.1$      & $\nu = 0.001$      &   $\epsilon=0.05$   & $\epsilon=0.01$   &  $\nu = 0.001$       &  $\nu = 0.0001$  \\ \midrule
 1000  & no      &  3.414e-2    &  6.532e-2   &  4.149e-2  & 1.050e-2  &  2.775e-2   & 1.545e-1   \\
 1000 & yes & \textbf{2.848e-2}  & \textbf{5.783e-2}  & \textbf{5.185e-3}   & \textbf{6.648e-3} &   \textbf{1.345e-2}  & \textbf{1.116e-1} \\ \midrule
 5000 & no      &  4.782e-3     &  1.571e-2     &  3.187e-3  & 7.872e-3 & 3.491e-3   & 9.450e-2\\
  5000 & yes  & \textbf{4.662e-3}     &  \textbf{1.257e-2} & \textbf{1.941e-3}   & \textbf{6.522e-3} &  \textbf{3.321e-3}    & \textbf{6.784e-2}   \\ \midrule
 10000 & no   &  3.460e-3    &  1.090e-2  &  2.945e-3 & 7.098e-3 &   2.626e-3  &  6.071e-2 \\
  10000&  yes &\textbf{2.436e-3}  & \textbf{9.561e-3}  & \textbf{9.388e-4} & \textbf{6.253e-3} &  \textbf{2.293e-3}    & \textbf{5.470e-2} \\
 \bottomrule
\end{tabular}}
\end{table}

  \begin{figure}
  \centering
	\subfigure{
	\begin{minipage}[b]{0.12\linewidth}
	\includegraphics[width=1\linewidth]{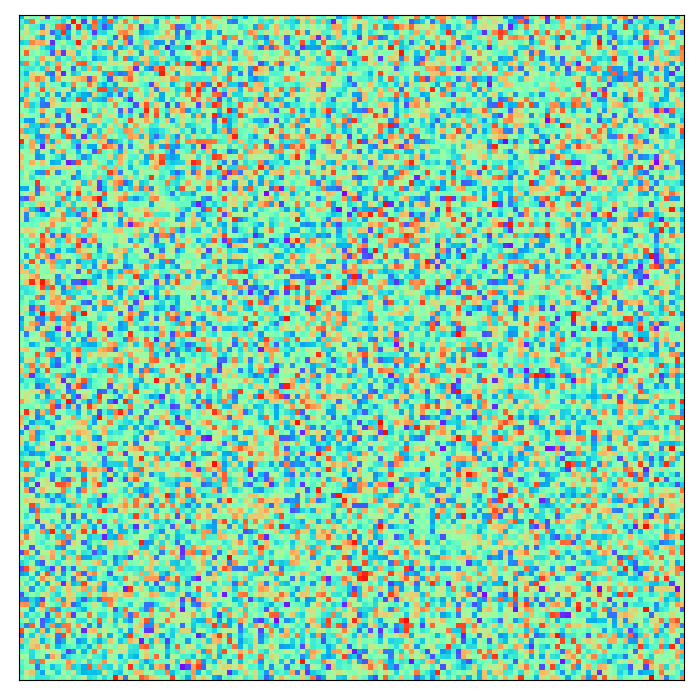}
	\includegraphics[width=1\linewidth]{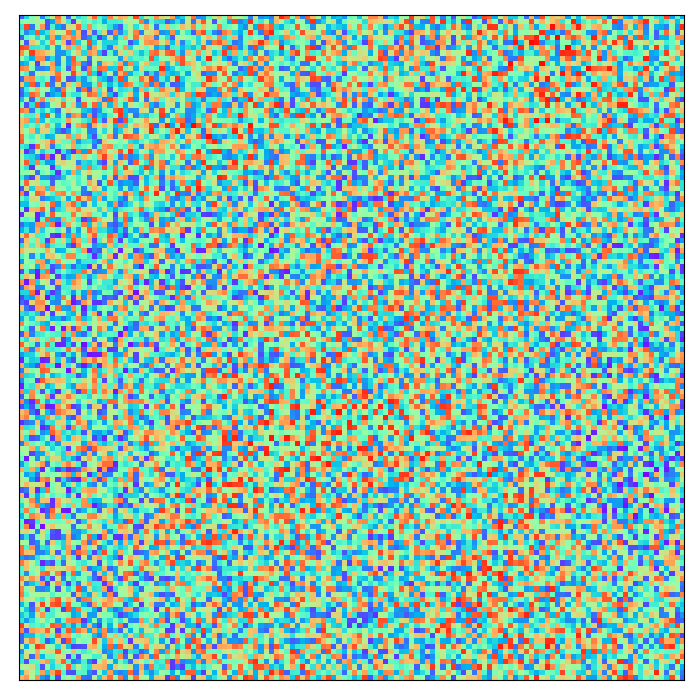}
	\includegraphics[width=1\linewidth]{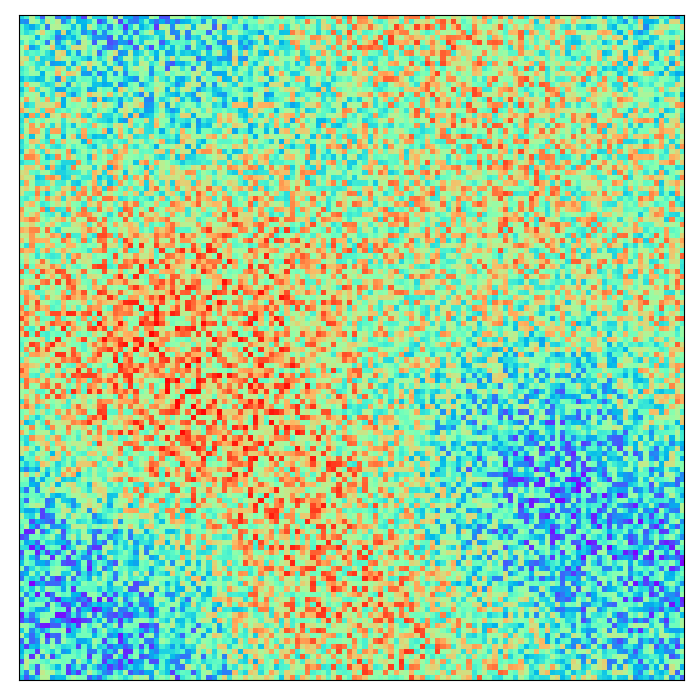}
	\centerline{(a)}
	\end{minipage}
	}\hspace{-2mm}
	\subfigure{
	\begin{minipage}[b]{0.12\linewidth}
	\includegraphics[width=1\linewidth]{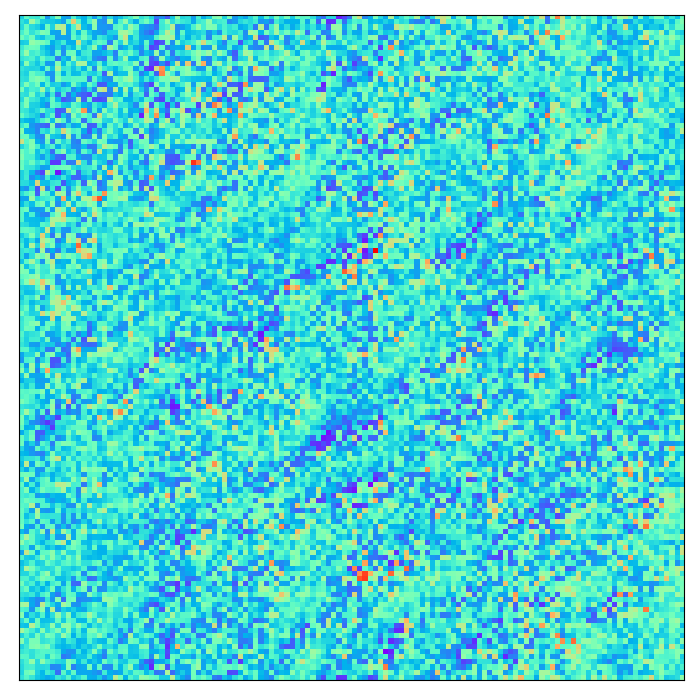}
	\includegraphics[width=1\linewidth]{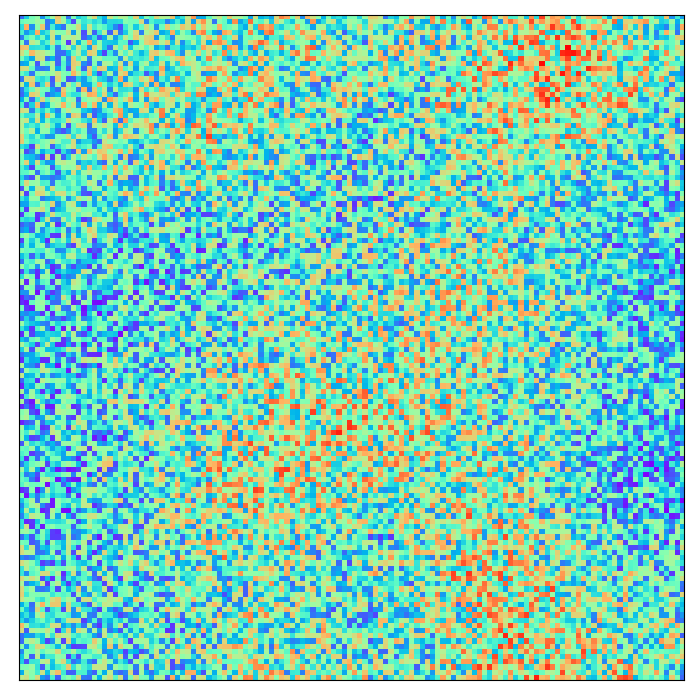}
	\includegraphics[width=1\linewidth]{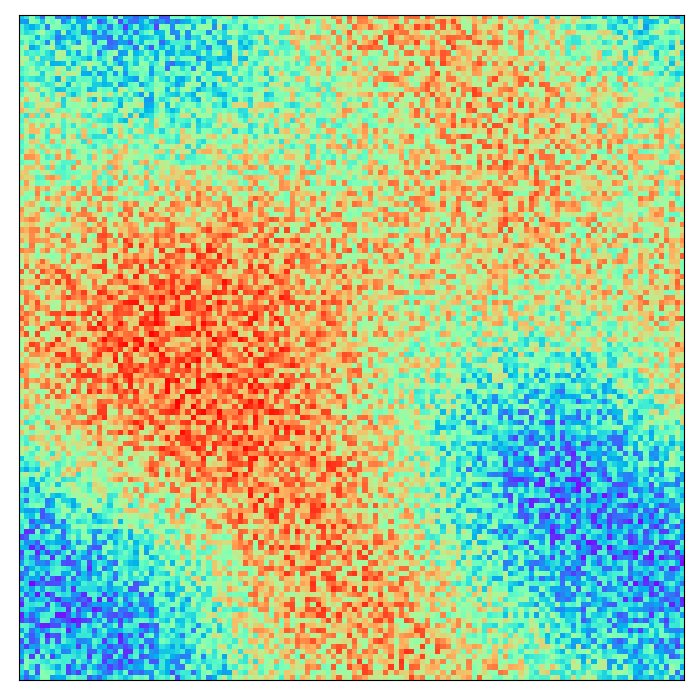}
	\centerline{(b)}
	\end{minipage}
	}\hspace{-2mm}
  	\subfigure{
	\begin{minipage}[b]{0.12\linewidth}
	\includegraphics[width=1\linewidth]{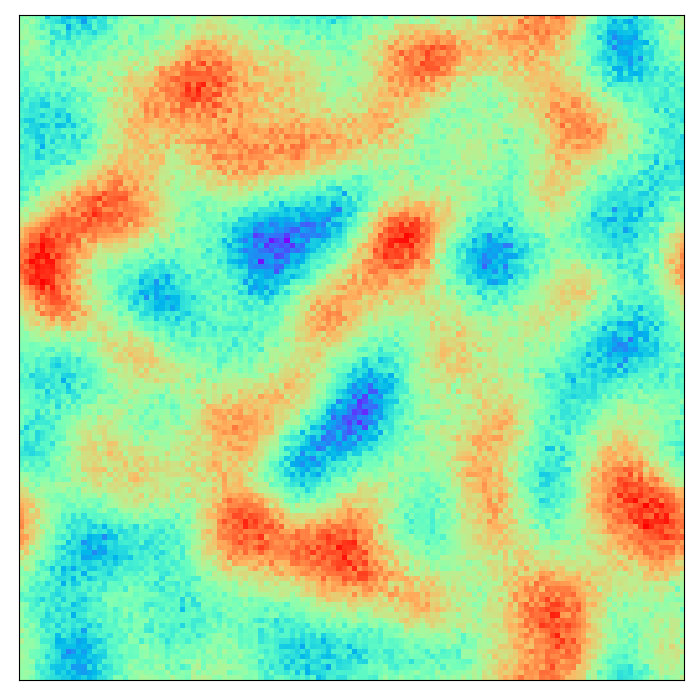}
	\includegraphics[width=1\linewidth]{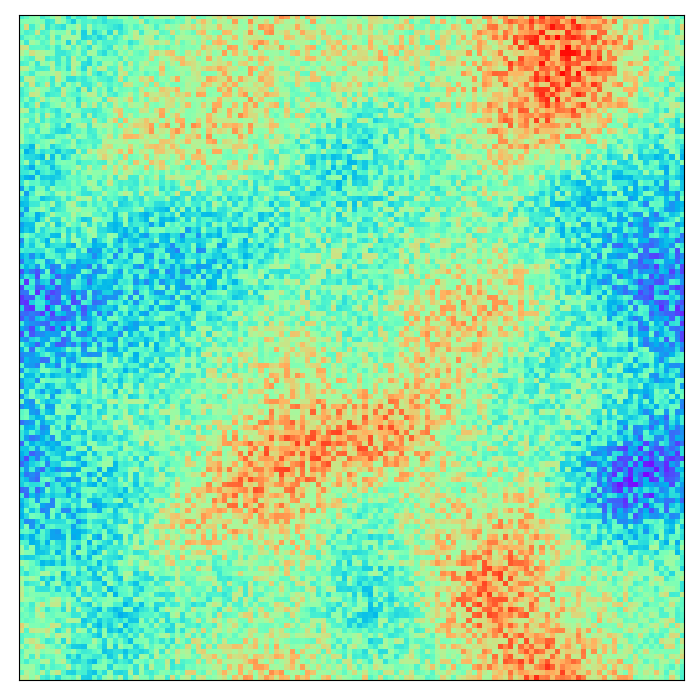}
	\includegraphics[width=1\linewidth]{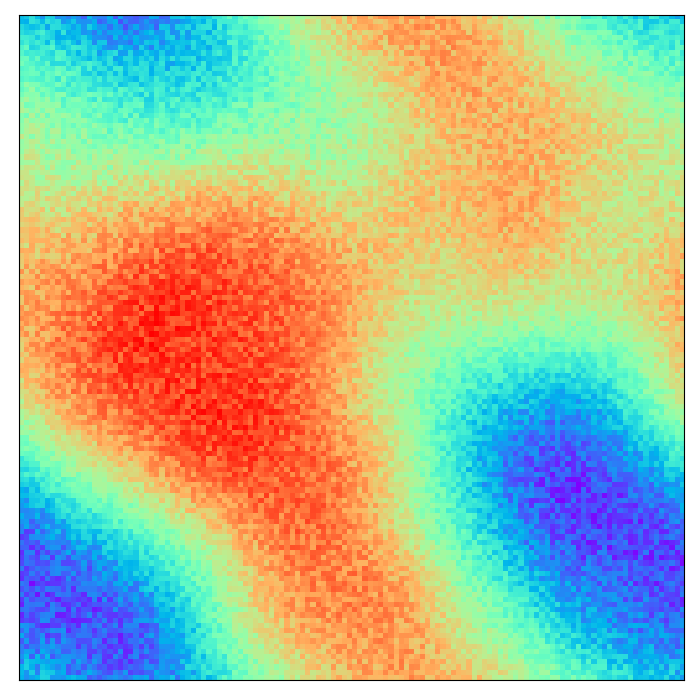}
	\centerline{(c)}
	\end{minipage}
	}\hspace{-2mm}
 	\subfigure{
	\begin{minipage}[b]{0.12\linewidth}
	\includegraphics[width=1\linewidth]{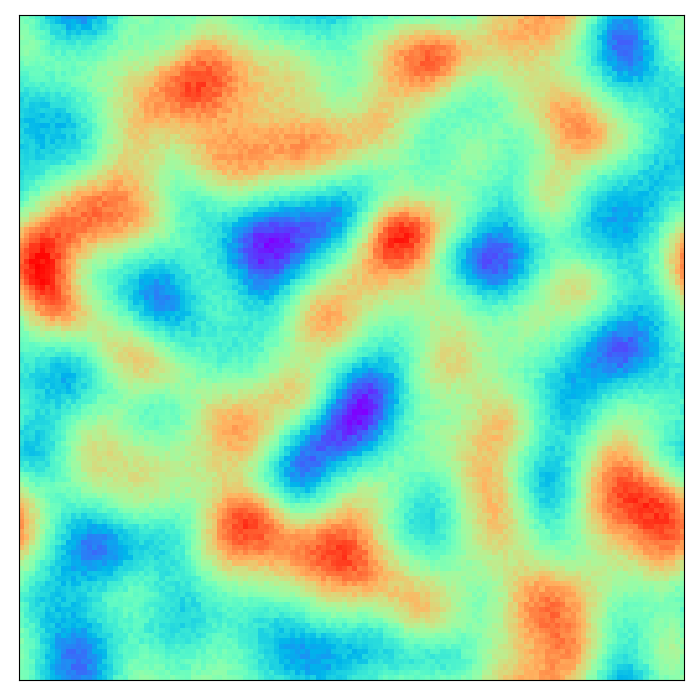}
	\includegraphics[width=1\linewidth]{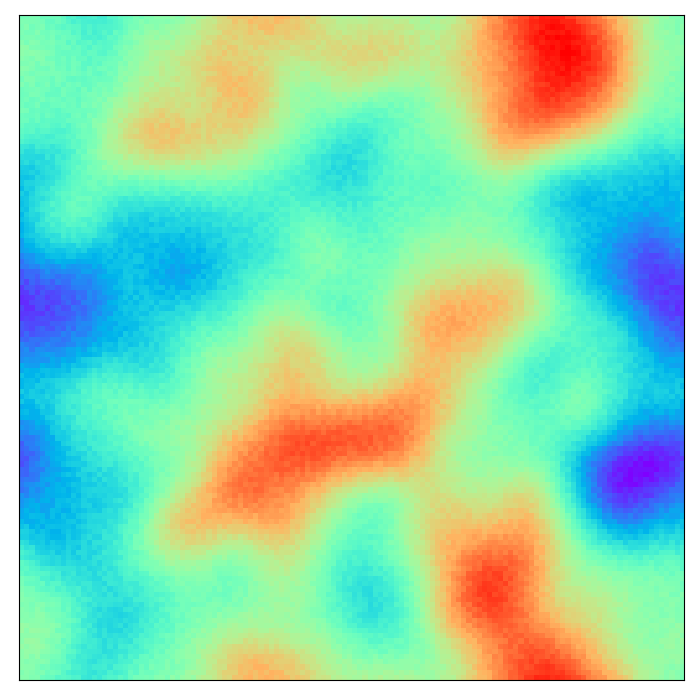}
	\includegraphics[width=1\linewidth]{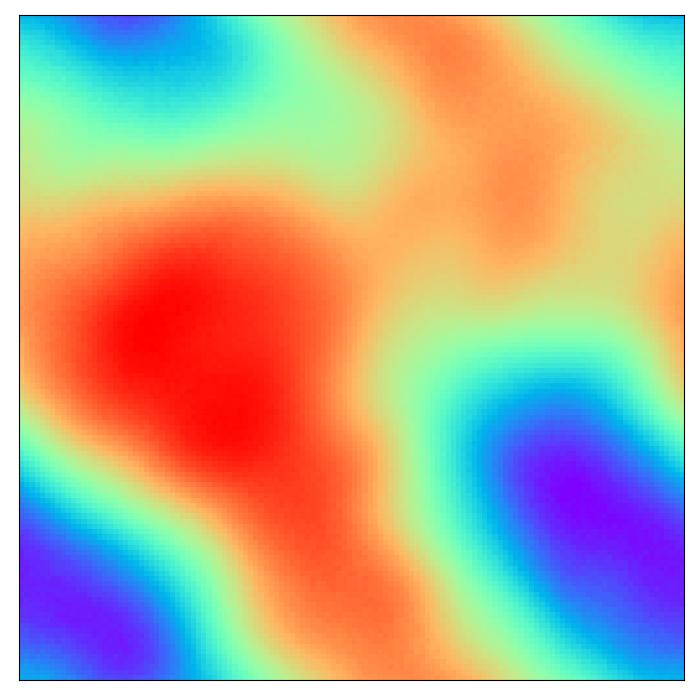}
	\centerline{(d)}
	\end{minipage}
	}\hspace{-2mm}
 	\subfigure{
	\begin{minipage}[b]{0.12\linewidth}
	\includegraphics[width=1\linewidth]{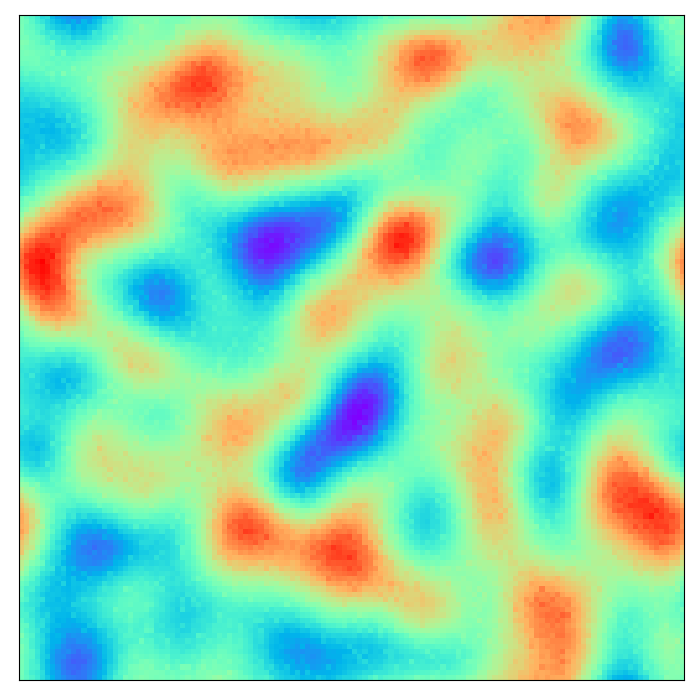}
	\includegraphics[width=1\linewidth]{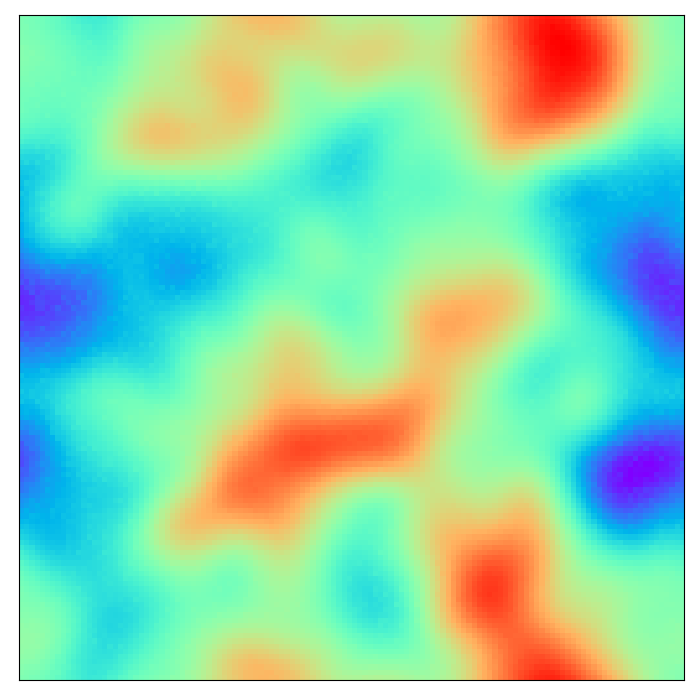}
	\includegraphics[width=1\linewidth]{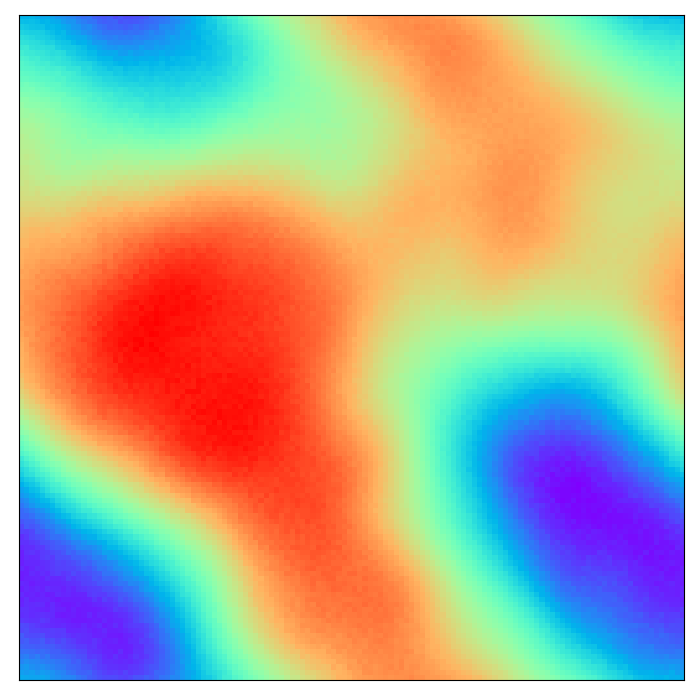}
	\centerline{(e)}
	\end{minipage}
	}\hspace{-2mm}
  	\subfigure{
	\begin{minipage}[b]{0.12\linewidth}
	\includegraphics[width=1\linewidth]{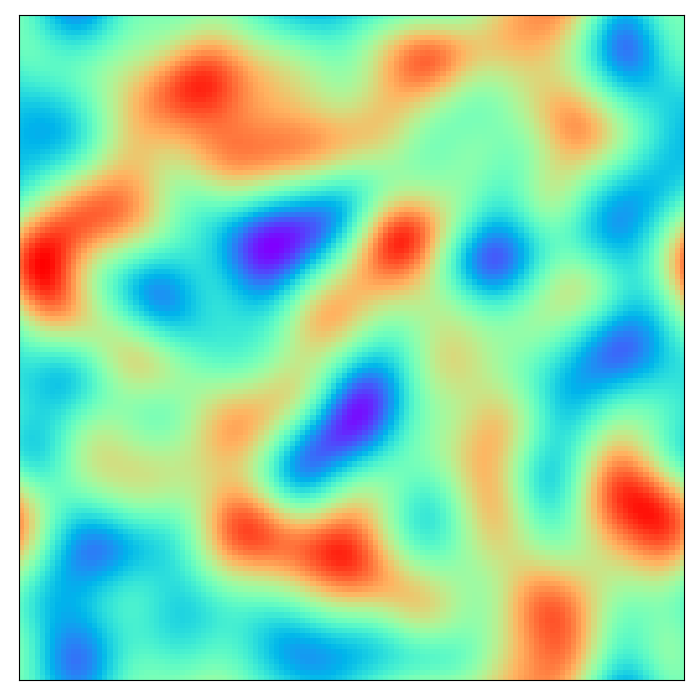}
	\includegraphics[width=1\linewidth]{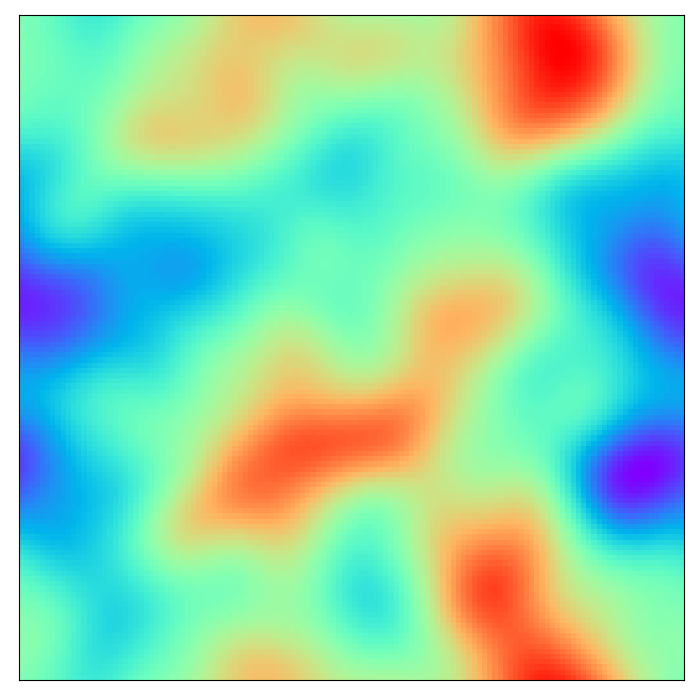}
	\includegraphics[width=1\linewidth]{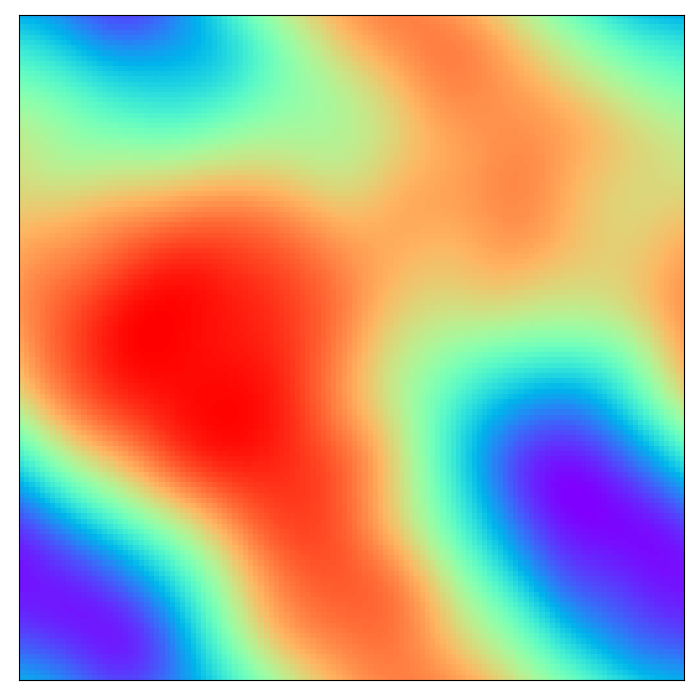}
	\centerline{(f)}
	\end{minipage}
	}
	\caption{Predictions of FNO trained on different datasets for Allen-Cahn equation ($\epsilon = 0.05$). (a) inputs with noise (b) 1000 data pairs (c) 10000 data pairs (d) 1000 data pairs + 1000 generated data pairs (e) 5000 data pairs + 5000 generated data pairs (f) ground truth}
 \label{fig:robustness}
  \end{figure}

\paragraph{Results} The test error for the three equations are shown in Table \ref{table_fno_iedata}. \modified{We compare the performance of FNO using both original training data and augmented training data. The original training data refers to the initially collected pairs of solutions, which serve as the baseline dataset. In contrast, the augmented training data includes both the original data pairs and additional pairs generated through inverse evolution data augmentation. Specifically, we use original datasets with 1000, 5000, and 10000 pairs, and the inverse evolution generates an equal number of additional pairs, resulting in augmented datasets of 2000, 10000, and 20000 pairs, respectively.} It can be observed that for all three equations, FNO on the augmented data have generally higher performance than that on the original data, especially when data size is small. In addition, we found in the experiments for the Allen-Cahn equation that the robustness of the FNO trained on the augmented data is also increased, which can be observed in Figure \ref{fig:robustness}. When noise is added to some regular input, the FNO trained on augmented data can still give a good prediction, while that trained on the original data fails. 

\subsection{UNet with Inverse Evolution Data Augmentation}
In addition to the FNO, we assess the impact of our data augmentation approach on the performance of the neural operator with a UNet architecture. Although UNet’s performance is generally inferior to FNO for solving PDEs, it is a well-established model across various fields, including scientific computing, and serves as a reliable baseline for evaluating the performance of neural operators. The experiments are conducted using the Burgers’ equation, the Allen-Cahn equation, and the Navier-Stokes equation. The results are presented in Table \ref{table_unet_iedata}. The data show that our augmentation method still enhances the performance of the UNet-based neural operator. However, the improvement is less pronounced compared to that observed with the FNO. In certain cases, the performance with augmented data is nearly equivalent to that obtained using only the original data. This may be attributed to the fact that the augmented data introduce more complex input structures. Consequently, neural operators with greater capacity to process complex inputs are likely to benefit more from our augmentation strategy, while those with lower capacity benefit less.

\begin{table}
    \centering
    \caption{Test relative error on $L^2$ norm for the UNet trained on different datasets. Again we see that our method is effective for the neural operators of the UNet architecture. }
    \label{table_unet_iedata}
    \resizebox{0.8\textwidth}{!}{\begin{tabular}{c|c|llllll}
    \toprule
     \multirow{2}{*}{Data Size}& \multirow{2}{*}{Augment.}& \multicolumn{2}{l}{Burgers' Equation} & \multicolumn{2}{l}{Allen-Cahn Equation} & \multicolumn{2}{l}{Navier-Stokes Equation} \\\cmidrule(lr){3-4}\cmidrule(lr){5-6}\cmidrule(lr){7-8}
     & & $\nu = 0.1$      & $\nu = 0.001$      &   $\epsilon=0.05$   & $\epsilon=0.01$   &  $\nu = 0.001$       &  $\nu = 0.0001$  \\ \midrule
     1000  & no      &  1.066e-1    &  1.035e-1   &  4.682e-2  & 1.349e-2  &  1.592e-2   & 1.139e-1   \\
     1000 & yes & \textbf{6.330e-2}  & \textbf{8.052e-2}  & \textbf{3.848e-2}   & \textbf{6.648e-3} &   \textbf{1.503e-2}  & \textbf{8.789e-2} \\ \midrule
     5000 & no      &  7.449e-3     &  4.260e-2     &  2.221e-2  & 8.219e-3 & 5.477e-3   & 5.304e-2\\
      5000 & yes  & \textbf{6.464e-3}     &  \textbf{3.214e-2} & \textbf{1.842e-2}   & \textbf{8.168e-3} &  \textbf{5.024e-3}    & \textbf{4.388e-2}   \\ \midrule
     10000 & no   &  6.443e-3    &  3.172e-2  &  1.691e-2 & \textbf{6.956e-3} &   3.179e-3  &  4.090e-2 \\
      10000&  yes &\textbf{3.543e-3}  & \textbf{2.706e-3}  & \textbf{1.595e-2} & 7.387e-3 &  \textbf{3.056e-3}    & \textbf{3.388e-2} \\
     \bottomrule
    \end{tabular}}
    \end{table}

\section{Discussion}
\label{conclusion}
\paragraph{Summary}This paper proposes a data augmentation technique based on inverse evolution for neural operators. This data generation method is easy to implement and can guarantee that the data obtained satisfies the implicit numerical schemes. We also propose high-order schemes for the inverse evolution to obtain data with higher accuracy. We carefully choose the initialization for inverse evolution to ensure the effectiveness of generated data. In addition, the instability caused by equations which have solutions with high gradients, is also considered in this paper and we introduced normalization and rescaling techniques to deal with them. We demonstrate the effectiveness of our data generation method by conducting a series of experiments on four different equations. Experimental results demonstrate that our data generation method indeed improve the performance of the neural operators and can also increase the robustness to noisy inputs for the Allen-Cahn equations. 

\paragraph{Limitations and Future Work}The proposed data augmentation approach is efficient and effective for most equations; however, the instability of inverse evolution remains a significant challenge. This issue is particularly undesirable when solving equations with chaotic solutions and evolving sharp interfaces. While the proposed preprocessing techniques can mitigate these instabilities, they also restrict the augmented data to a specific space rather than the original solution space. Additionally, instability poses challenges for large time steps. To address the instability issue, designing new numerical schemes such as appropriate adaptive methods for inverse evolution is a viable solution. In this paper, we focus on explicit methods for inverse evolution because they ensure that the obtained data align with corresponding implicit methods. However, exploring high-order semi-implicit methods for inverse evolution could enhance stability while maintaining accuracy. Furthermore, initialization is fundamental to our data augmentation method. It is crucial to investigate how to generate appropriate initial values to ensure the generated data better covers the solution space. Spatial discretization methods also play a pivotal role for inverse evolution. Implementing suitable spatial discretization methods can mitigate instability, thus enabling more flexible initialization for inverse evolution.  

\bibliographystyle{plain} 
\bibliography{ie_for_fno.bib}

\end{document}